\documentclass[10pt,journal]{IEEEtran}

\PassOptionsToPackage{table}{xcolor}

\ifCLASSOPTIONcompsoc
  \usepackage[nocompress]{cite}
\else
  \usepackage{cite}
\fi
\usepackage{subfloat}
\usepackage{subfig}
\usepackage{graphicx}
\usepackage{xcolor}
\usepackage{amsmath}
\usepackage[ruled,linesnumbered]{algorithm2e}
\usepackage{booktabs}
\usepackage{multirow}
\usepackage{makecell}
\usepackage{amssymb}
\usepackage{lipsum}
\usepackage{colortbl}
\usepackage{soul}
\usepackage{microtype}
\usepackage{stfloats}

\usepackage[citecolor=green, colorlinks]{hyperref}
\definecolor{whitesmoke}{rgb}{0.96, 0.96, 0.96}

\newcommand{\revise}[1]{\textcolor{black}{#1}}
\newcommand{\major}[1]{\textcolor{black}{#1}}
\newcommand{\minor}[1]{\textcolor{black}{#1}}

\begin{document}
\title{Self-supervised Learning of LiDAR 3D Point Clouds via 2D-3D Neural Calibration}

%
%
%
%

\author{Yifan Zhang, 
        Junhui Hou,~\IEEEmembership{Senior Member,~IEEE},
        Siyu Ren, 
        Jinjian Wu,~\IEEEmembership{Senior Member,~IEEE},
        Yixuan Yuan,~\IEEEmembership{Senior Member,~IEEE},
        and Guangming Shi,~\IEEEmembership{Fellow,~IEEE}
\IEEEcompsocitemizethanks{\IEEEcompsocthanksitem Y. Zhang is with the School of Mechatronic Engineering and Automation, Shanghai University, Shanghai, China, and also with the Department of Computer Science, City University of Hong Kong, Hong Kong. E-mail: yfzhang@shu.edu.cn;
\IEEEcompsocthanksitem J. Hou, and S. Ren are with the Department of Computer Science, City University of Hong Kong, Hong Kong. E-mail: siyuren2-c@my.cityu.edu.hk; jh.hou@cityu.edu.hk;
\IEEEcompsocthanksitem J. Wu and G. Shi are with the School of Artificial Intelligence, Xidian University,
Xi’an 710071, China (e-mail: jinjian.wu@mail.xidian.edu.cn; gmshi@xidian.edu.cn)
\IEEEcompsocthanksitem Y. Yuan is with the Department of Electronic Engineering, The Chinese University of Hong Kong, Hong Kong. E-mail: yxyuan@ee.cuhk.edu.hk
\IEEEcompsocthanksitem  
This work was supported in part by the NSFC Excellent Young Scientists Fund 62422118, in part by the Hong Kong RGC under Grants 11219324 and 11219422, and in part by the Hong Kong ITC under Grant ITS/164/23. (\textit{Corresponding author: Junhui Hou})
 \protect \\

}
}

%
%

\markboth{}%
{Shell \MakeLowercase{\textit{et al.}}: Bare Demo of IEEEtran.cls for Computer Society Journals}
%



\IEEEtitleabstractindextext{%
\begin{abstract}
This paper introduces a novel self-supervised learning framework for enhancing 3D perception in autonomous driving scenes. Specifically, our approach, namely NCLR, focuses on 2D-3D neural calibration, a novel pretext task that estimates the \revise{rigid pose} aligning camera and LiDAR coordinate systems. First, we propose the learnable transformation alignment to bridge the domain gap between image and point cloud data, converting features into a unified representation space for effective comparison and matching. Second, we identify the overlapping area between the image and point cloud with the fused features. Third, we establish dense 2D-3D correspondences to estimate the \revise{rigid pose}. The framework not only learns fine-grained matching from points to pixels but also achieves alignment of the image and point cloud at a holistic level, understanding \minor{the LiDAR-to-camera extrinsic parameters}.
 We demonstrate the efficacy of NCLR by applying the pre-trained backbone to downstream tasks, such as LiDAR-based 3D semantic segmentation, object detection, and panoptic segmentation. Comprehensive experiments on various datasets illustrate the superiority of NCLR over existing self-supervised methods. The results confirm that joint learning from different modalities significantly enhances the network's understanding abilities and effectiveness of learned representation. 
The code is publicly available at \url{https://github.com/Eaphan/NCLR}.
\end{abstract}

}

\maketitle

\IEEEdisplaynontitleabstractindextext

%
\IEEEpeerreviewmaketitle

\section{Introduction}\label{sec:introduction}

\IEEEPARstart{T}{he} LiDAR technology serves as a vital enhancement to 2D cameras by precisely capturing the surroundings of a vehicle, offering robust performance in challenging conditions, including low light, intense sunlight, or glare from approaching headlights. This 3D perception, derived from LiDAR point clouds, is essential for the effective navigation of autonomous vehicles in three-dimensional spaces. While current leading methods are based on extensive labeled datasets, labeling 3D data is exceedingly expensive and time-intensive, given the limited availability of annotation resources. Consequently, there is a growing imperative to utilize unlabeled data. This approach aims to minimize the need for extensive annotation while enhancing the effectiveness of networks trained on a limited amount of labeled data.

An effective strategy to tackle this challenge involves initially pre-training a neural network solely on unannotated data, for example, by employing a pretext task that eliminates the need for manual labeling. Subsequently, this self-supervised pre-trained network can be fine-tuned for specific downstream tasks. Through thorough pre-training, the network acquires initial weights that serve as a beneficial foundation for additional supervised training. As a result, training the network for a particular downstream task generally demands fewer annotations to achieve comparable performance levels to those of a network trained from scratch. The self-supervised approaches have been very successful in 2D (images), even reaching the level of supervised pre-training~\cite{jing2020self,liu2021self}.

A number of self-supervised approaches have also been successful in the field of point cloud data processing. 
Several methods explore different levels of representation (point-level~\cite{xie2020pointcontrast,chen2022_4dcontrast}, segment-level~\cite{nunes2022segcontrast}, region-level~\cite{liang2021gcc3d,yin2022proposalcontrast}) and introduce contrastive losses to capture the geometric and structural nuances of 3D data~\cite{zhang2021depthcontrast,sautier2023bevcontrast}. 
Another class of methods takes the temporal correlation as a form of supervision~\cite{nunes2023TARL,huang2021STRL,wu2023STSSL}. For example, STRL~\cite{huang2021STRL} processes two adjacent frames from 3D point cloud sequences. It transforms the input with spatial data augmentation and learns an invariant representation.
And TARL~\cite{nunes2023TARL} exploit vehicle movement to match objects in consecutive scans. It then trains a model to maximize the segment-level feature similarities of the associated object in different scans, enabling the learning of a consistent representation across time.
Reconstruction-based methods~\cite{pang2022masked,zhang2022pointm2ae} apply masked auto-encoding to point cloud and reconstructed points coordinates using the Chamfer distance. ALSO~\cite{boulch2023ALSO} proposes to use unsupervised surface reconstruction as a pretext task to train 3D backbones. 

\begin{figure}[t]
	\centering
	\includegraphics[width=0.48\textwidth]{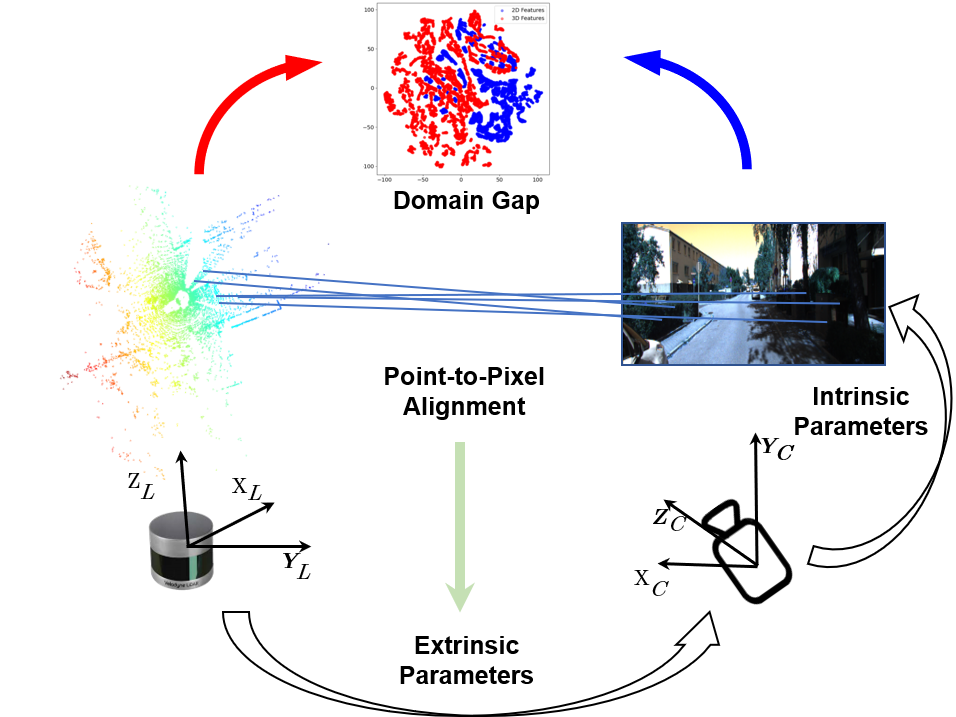} 
	\caption{
		Illustration of the proposed 2D-3D neural calibration developed as the pretext task for self-supervised learning of 3D LiDAR point clouds. Our method not only learns the local matching from points to pixels via contrastive losses but also estimates the (unknown) holistic \revise{rigid pose} aligning the camera and LiDAR systems. Besides, we propose a learnable transformation alignment to fill the domain gap between the image and the point cloud. 
	}
	\label{fig:pretext_task}
\end{figure}

Although multi-modal self-supervised learning holds significant promise for point cloud applications, its potential has not been completely realized. Current approaches predominantly utilize unlabeled data from synchronized and calibrated camera-LiDAR setups to pretrain 3D backbones. The core strategy involves identifying corresponding point-pixel pairs and ensuring their representations closely align~\cite{li2022simipu,sautier2022image,mahmoud2023self}. However, these methods face two primary limitations: 
 First, existing image-to-point self-supervised learning methods only align the pixels and points locally through contrastive learning, but ignore the holistic spatial relationship between the image and point cloud. Second, these methods neglect the inherent differences in the characteristics of these distinct modalities when aligning the corresponding features of pixels and points. As a result, their performance is still limited.

In this paper, we introduce a novel pretext task, the 2D-3D neural calibration, for the self-supervised pre-training of networks for 3D perception in autonomous driving scenes, as shown in Fig.~\ref{fig:pretext_task}. 
During pre-training, our pretext task not only learns fine-grained matching from individual points to pixels but also achieves a comprehensive alignment between the image and point cloud data, i.e., understanding \minor{the LiDAR-to-camera extrinsic parameters}. Specifically, to impose supervision on the \revise{rigid pose} estimation, we propose an end-to-end differentiable framework distinguished by its integration of a soft-matching strategy and a differentiable PnP solver. We posit that this joint learning from different modalities for both local and global-level alignment will enhance the network's capacity for sophisticated understanding and enable it to develop effective representations.
 In addition, to fill the substantial domain gap between point clouds and images, we propose a learnable transformation alignment during pre-training, replacing the direct cosine distance alignment. This method converts features into a unified representation space, allowing for more accurate feature comparison and matching beyond the capabilities of cosine distance.

\begin{figure}[t]
	\centering
	\includegraphics[width=0.48\textwidth]{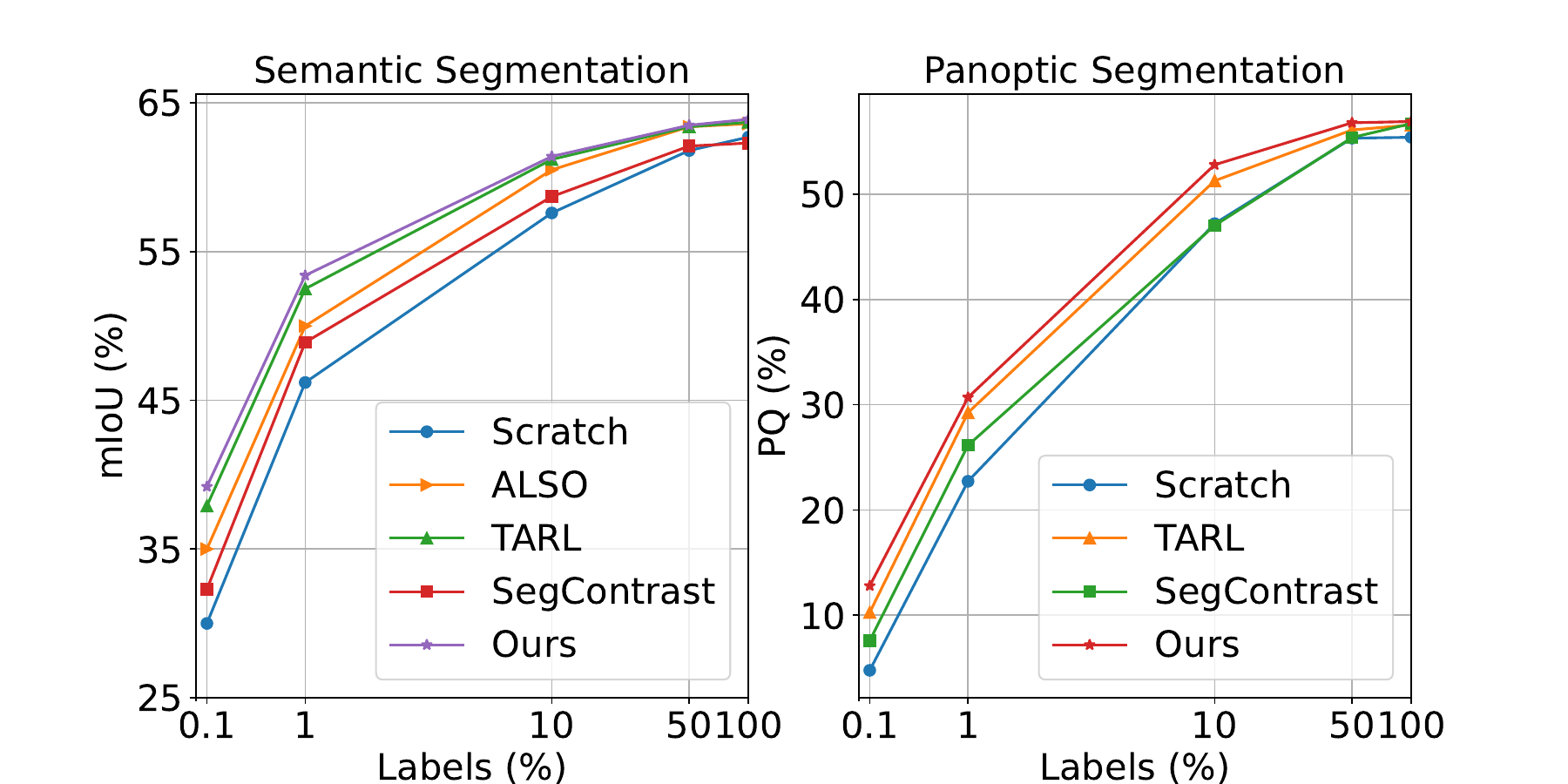} 
	\caption{
		\minor{Semantic and panoptic segmentation performance comparison of our method against existing 3D self-supervised learning methods and scratch training on the SemanticKITTI dataset. Our approach achieves comparable results with only 10\% annotated data, demonstrating near-equivalent effectiveness to full dataset training from scratch.}
	}
        \vspace{-0.2cm}
	\label{fig:seg_performance}
\end{figure}

To evaluate our method, we carry out comprehensive experiments and compare it against state-of-the-art studies on downstream tasks, including 3D semantic segmentation, object detection, and panoptic segmentation. The results demonstrate that our method outperforms existing self-supervised learning approaches, as evidenced by its superior performance in adapting to various downstream tasks and datasets~\cite{behley2019semantickitti,Geiger_KITTI,caesar2020nuscenes,pan2020semanticposs}. 
Fig.~\ref{fig:seg_performance} shows that our pre-training method enables the network to reach performance comparable to scratch training with fewer annotations for downstream tasks.


To summarize, the main contributions of this work are as follows:
\begin{itemize}
    \item We introduce an innovative perspective to self-supervised learning, centered on achieving a thorough alignment between two distinct modalities.
    \item We identify the inherent domain gap between the image and the point cloud and propose a learnable transformation alignment for feature comparison.
    \item We propose an end-to-end network featuring the soft-matching strategy and a differentiable PnP solver, achieving state-of-the-art performance on the 2D-3D neural calibration task.
    \item We demonstrate the superiority of our method over existing self-supervised learning methods across three distinct downstream tasks in 3D perception.
\end{itemize}

The rest of this paper is organized as follows. Section~\ref{sec:related_work} provides an overview of related literature pertinent to our research. In Section~\ref{sec:method}, we present the overall architecture of NCLR and elaborate its principal elements. Section~\ref{sec:experiments} presents an empirical evaluation of our proposed approach across three distinct downstream tasks, along with ablation studies to assess the impact of key components. The paper concludes with Section~\ref{sec:conclusion}, summarizing our findings.

\section{Related Work}\label{sec:related_work}

\subsection{Self-supervised Learning for Images}
Self-supervised learning involves training models on unlabeled data by generating pseudo-labels or tasks based on the data itself, without human annotations. The model learns representations by solving pretext problems derived from the inherent structure of the unlabeled inputs.
The field of self-supervised learning on images has seen significant evolution over the years. Initial approaches are centered around pretext tasks, which are auxiliary tasks designed to derive meaningful data representations. 
These tasks involve teaching models to restore color in black-and-white images~\cite{zhang2016colorful} and to ascertain the spatial relation between image segments~\cite{doersch2015unsupervised}.

The advent of deep learning has brought about a paradigm shift in the field. One such approach is contrastive learning~\cite{khosla2020supervised,chen2020simple}. 
Contrastive learning involves training networks to distinguish between different augmentations of the same image versus those from distinct images. It minimizes the representation distance for augmentations of the same image (positive pairs) while maximizing it for different images (negative pairs).
Generative models~\cite{liu2021self,donahue2019large} have also emerged as a significant trend in self-supervised learning. These models, often employing Variational Autoencoders (VAEs) or Generative Adversarial Networks (GANs), learn to generate images that closely resemble the training data. By learning to generate images, these models capture the underlying data distribution, which can be leveraged for various downstream tasks.
In addition to these, there have been several exciting advancements in the field. Masked image modeling, for instance, involves masking parts of an image and training a model to predict the masked parts~\cite{he2022masked}. Multi-modal models aim to learn representations across different modalities (e.g., text, image, audio) using self-supervised learning. The Vision Transformer (ViT) adapts the transformer architecture, initially developed for natural language processing, to computer vision tasks~\cite{caron2021emerging}. Additionally, there has been explored into incorporating physics-based priors in self-supervised learning.


\subsection{Self-supervised Learning for Point Clouds}
The advances in self-supervision on point clouds have closely followed the improvements made in images. Early self-supervised methods use pretext tasks such as predicting transformations applied to the point cloud or reconstructing parts of the point cloud~\cite{poursaeed2020self,sauder2019self}. These methods are applied on dense scans of single objects~\cite{zhang2023pointvst,zhang2023pointmcd}.

\vspace{0.5em}
\noindent\textbf{Discriminative-based Methods.} This type of method explores different levels of representation (point-level~\cite{xie2020pointcontrast,chen2022_4dcontrast}, segment-level~\cite{nunes2022segcontrast}, region-level~\cite{liang2021gcc3d,yin2022proposalcontrast}) and introduces contrastive losses to capture the geometric and structural nuances of 3D data~\cite{zhang2021depthcontrast,sautier2023bevcontrast}. Nunes et al. introduced a novel contrastive learning approach focused on understanding scene structure. This method involves extracting class-agnostic segments from point clouds and employing contrastive loss on these segments to differentiate between structurally similar and dissimilar elements~\cite{nunes2022segcontrast}.

\vspace{1em}
\noindent\textbf{Temporal-consistency-based Methods.} 
STRL~\cite{huang2021STRL} utilizes pairs of temporally-linked frames from 3D point cloud sequences, applying spatial data augmentation to learn invariant representations. STSSL~\cite{wu2023STSSL} incorporates spatial-temporal positive pairs, introducing a point-to-cluster technique for spatial object distinction and a cluster-to-cluster method using unsupervised tracking for temporal correlations. TARL~\cite{nunes2023TARL} uses vehicle motion to align objects over time in different scans, focusing on enhancing segment-level feature similarities for consistent temporal representations. 
These works demonstrate how leveraging temporal information can lead to robust and invariant representations, which are beneficial for various downstream tasks.

\vspace{0.5em}
\noindent\textbf{Reconstruction-based Methods.} Reconstruction-based methods have also been successful for self-supervision on point clouds. Some methods reconstruct point coordinates using the Chamfer distance~\cite{pang2022masked,zhang2022pointm2ae}.
Point-BERT randomly masks some patches and feeds them into Transformers to recover the original tokens at the masked locations~\cite{yu2022pointbert}.
Recently, ALSO~\cite{boulch2023ALSO} proposes to use unsupervised surface reconstruction as a pretext task to train 3D backbones on automotive LiDAR point clouds. Using the knowledge of occupancy before and after an observed 3D point along a LiDAR ray, it learns to construct an implicit occupancy function and good point features.
\revise{Besides, rendering-based techniques~\cite{yang2023unipad,huang2023ponder,zhu2023ponderv2} reconstruct masked scenes from point clouds or multi-view images, which have achieved promising generalization ability.}

\begin{figure*}[t]
	\centering
	\includegraphics[width=0.95\textwidth]{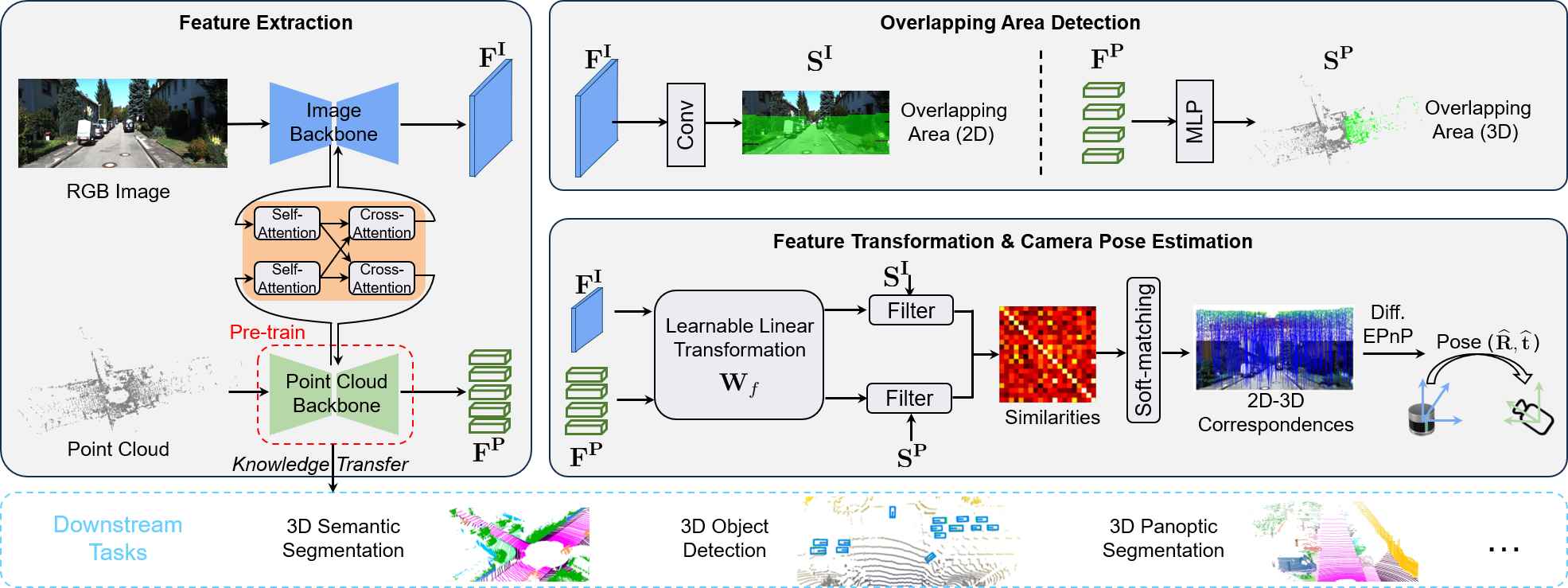} 
	\caption{
		\major{The overall pipeline for 2D-3D neural calibration serves as a pretext task for generating self-supervised representation for LiDAR 3D point cloud data. 
     This process involves four key steps: (1) Feature extraction from both the image and point cloud. (2) Detection of overlapping areas using fused features from both modalities. (3) Transformation of image and point cloud features into a unified representation space for similarity computation. We filter the points and pixels in estimated non-overlapping areas. (4) Application of a soft-matching strategy to establish 2D-3D correspondences, followed by the use of a differentiable EPnP solver for camera pose estimation. }
	}
	\label{fig:pipeline}
\end{figure*}

\vspace{0.5em}
\noindent\textbf{Multi-modal Self-supervised Learning.}
\revise{Another line of work leverages synchronized and calibrated cameras and LiDAR to pretrain a 3D backbone~\cite{li2022simipu,sautier2022image,mahmoud2023self}. These contrastive methods find pairs of corresponding points and pixels and ensure that the associated point and pixel representations are as close as possible~\cite{gilles2024three}.
The primary motivation of these methods is to transfer the knowledge already embedded in a pre-trained 2D backbone to the 3D backbone. In contrast, our approach trains the network to achieve holistic alignment between the image and point cloud, enabling it to understand the two modalities thoroughly and subsequently learn effective 3D representations. Besides, our method does not rely on a pre-trained 2D backbone. By eliminating this dependency, our approach introduces greater flexibility and removes the constraint of inheriting specific representations from a 2D network.}

\subsection{LiDAR-based 3D Perception} 
LiDAR-based 3D perception is a critical component in many applications, particularly in autonomous driving. 
It encompasses several tasks, including 3D semantic segmentation, 3D object detection, and 3D panoptic segmentation.
Due to space limitations, this paper only discusses the methods we employed. For additional methodologies, we refer readers to the review paper~\cite{guo2020deep}.

\vspace{0.5em}
\noindent\textbf{3D Semantic Segmentation.} One of the foundational works in this area is PointNet~\cite{qi2017pointnet}, which marks a significant departure from previous methods by directly processing point clouds for tasks like segmentation. 
MinkUNet adapts the U-Net architecture for sparse 3D point clouds, using Minkowski Engine for efficient segmentation~\cite{choy20194d}. SPVCNN combines sparse and continuous convolutions for scalable and efficient 3D point cloud segmentation~\cite{tang2020searching}.

\vspace{0.5em}
\noindent\textbf{3D Object Detection.} 
Advancements in 3D object detection have led to its categorization into voxel-based, point-based, and hybrid methods~\cite{zhang2023glenet}. Voxel-based techniques convert point clouds into uniform grid structures for feature extraction using sparse convolutions~\cite{zhou2018voxelnet,yan2018second}. Point-based methods directly handle raw 3D point clouds to make predictions. Hybrid models, such as PV-RCNN~\cite{shi2020pv} and PDV~\cite{hu2022point}, merge the advantages of both voxel and point-based approaches, optimizing both accuracy and efficiency in detection.

\vspace{0.5em}
\noindent\textbf{3D Panoptic Segmentation.} 
\major{The task of 3D panoptic segmentation requires understanding both objects and scenes. Hong et al. \cite{hong2021lidar} introduced the Dynamic Shifting Network (DS-Net), using cylinder convolution for LiDAR data and a bottom-up clustering approach for instance separation. Recently, Marcuzzi et al.\cite{marcuzzi2023mask} proposed an end-to-end 3D panoptic segmentation approach using query-based transformers to predict binary masks with semantic categories, eliminating the need for post-processing clustering~\cite{marcuzzi2023ralmeem}. 
Yilmaz et al. ~\cite{yilmaz2024mask4former} proposed the first transformer-based approach for 4D panoptic segmentation.
}

\section{Proposed Method}\label{sec:method} 
\subsection{Overview}\label{sec:overview} 
We propose a novel approach to generating self-supervised features for 3D point clouds, utilizing the 2D-3D neural calibration as a pretext task. 
The 2D-3D neural calibration aims to identify the unknown \revise{rigid pose}, comprising a rotation $\mathbf{R} \in SO(3)$ and a translation $\mathbf{t}\in \mathbb{R}^3$, through a neural network. This pose aligns the coordinate systems of the camera and the point cloud.
We denote the input image as $\mathbf{I} \in \mathbb{R}^{3\times H\times W}$ and point cloud as $\mathbf{P}=\{\mathbf{p}_{i}\in \mathbb{R}^3|i=1,2,...,N\}$, where $H$ and $W$ are the height and width of the image, and $N$ is the number of points. 
To achieve effective representations, our pre-training method not only learns fine-grained matching from points to pixels but also achieves alignment of the image and point cloud at a holistic level, i.e., understanding~\minor{the LiDAR-to-camera extrinsic parameters}.
Then the point cloud backbone could be used for downstream 3D perception tasks.

Fig.~\ref{fig:pipeline} illustrates our overall framework. Initially, we extract features from both the image and point cloud using dedicated backbones. Next, we identify overlapping regions in both the image and point cloud, leveraging the fused features, as overlaps between them are partial. Subsequently, a learnable transformation is applied to harmonize the image and point cloud features into a single representation space before similarity computations between them. The final step involves employing a soft-matching approach to establish 2D-3D correspondences. These correspondences are inputted into a differentiable EPnP solver for camera pose estimation. The proposed framework is designed to be end-to-end trainable.


\subsection{Feature Extraction}\label{sec:feature_extraction}

Both the image and point cloud branches employ an encoder-decoder structure. After the encoder stage, we fuse the downsampled 2D feature map and sampled keypoint features to combine these diverse features into a unified representation.

\vspace{0.5em}
\noindent\textbf{Transformer-based Feature Fusion}. We perform bidirectional feature fusion based on the attention mechanism. We denote the downsampled 2D feature map after encoder as $\mathbf{F}_E^\mathbf{I} \in \mathbb{R}^{H_{E}\times W_{E}\times C_{E}} $ and features of sampled keypoints as $\mathbf{F}_E^\mathbf{P} \in \mathbb{R}^{N_{E}\times C_{E}}$. 
Each fusion layer comprises three components: first, a multi-head self-attention layer for image and point features; second, a multi-head cross-attention layer that refines each domain's features using data from the other; and third, a point-wise feed-forward network. These cross-attention layers facilitate the model's understanding of interrelations and complementary aspects across diverse data types.
To incorporate the positional information, we add sinusoidal position encoding to the inputs of transformer layers~\cite{vaswani2017attention}. Then the fused features are passed to the decoders to obtain higher-resolution feature maps. The outputs of decoder are features $\mathbf{F}^\mathbf{I} \in \mathbb{R}^{H'\times W'\times C}$ and $\mathbf{F}^\mathbf{P} \in \mathbb{R}^{N \times C}$.

Note that our framework is effective for not only point-based but also voxel-based backbones for the point cloud branch. If we need to pre-train a voxel-based backbone for downstream tasks, we can incorporate it and obtain the final point features by aggregating the voxel-wise features and fine-grained point-wise features.

\subsection{Feature Discrimination}

\begin{figure}[t]
	\centering
	\includegraphics[width=0.48\textwidth]{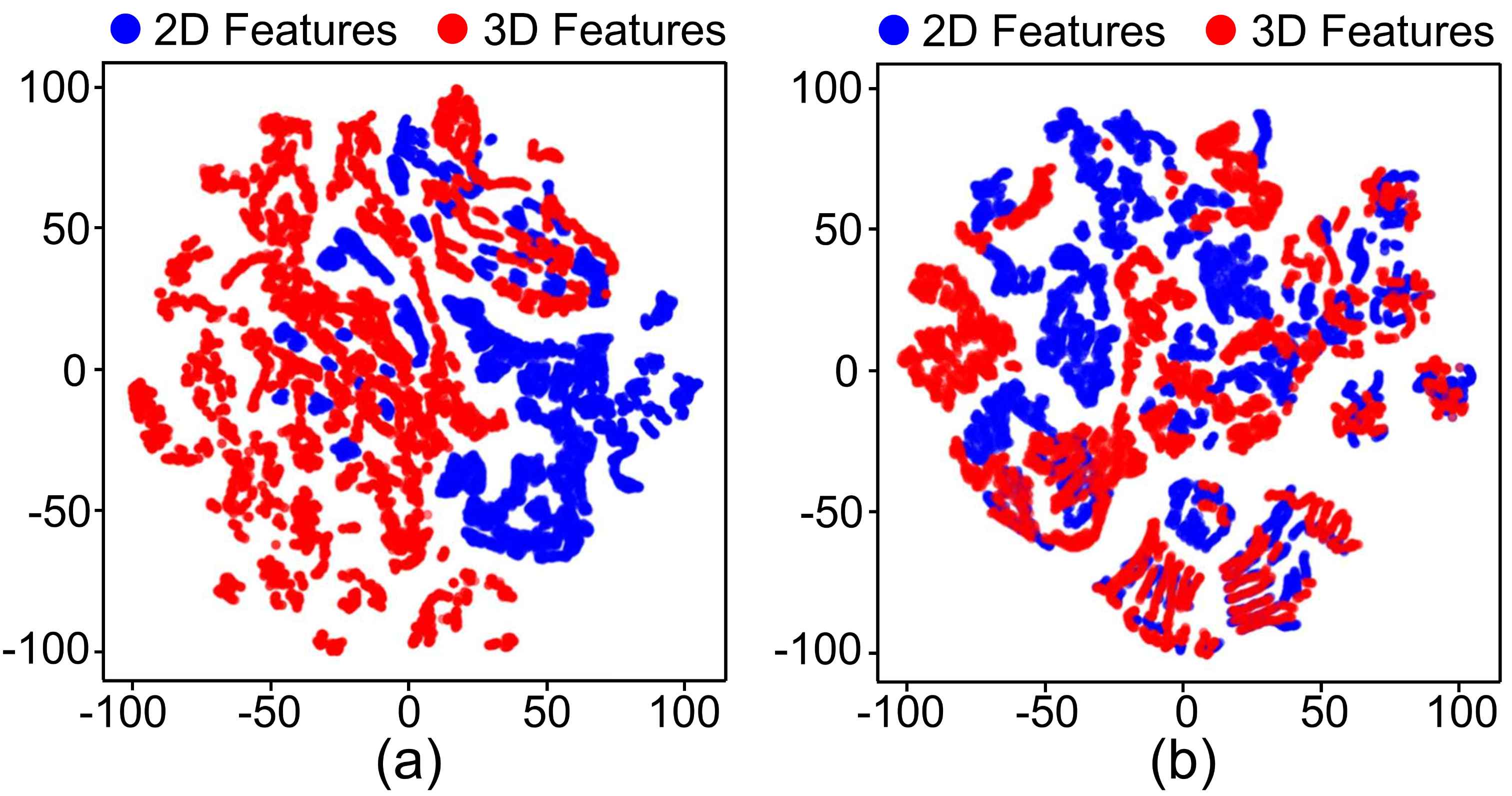} 
	\caption{
	\revise{Visual illustration of t-SNE clustering for image and point cloud features as learned by the model, shown (a) without and (b) with the incorporation of learnable transformation alignment.}
	}
	\label{fig:feature_tsne}
\end{figure}

Building dense point-to-pixel correspondences relies on learning discriminative features. 
Previous work~\cite{ren2022corri2p} simply uses the \textit{cosine distance} as descriptor loss to bring point and image features of positive pairs closer and those of negative pairs farther apart. The loss function drives the matching features to be similar. Although such a method tries to minimize the distribution gaps between the two different modalities by feature fusion, it is impossible to fully eliminate such discrepancy. 

To align the two modalities, we apply an InfoNCE~\cite{oord2018representation} loss on the fused features. 
Specifically, we sample a subset of points $\mathbf{P}_S$ inside the overlapped areas with the RGB image, then we can compute the InfoNCE loss for these points as:
\begin{equation}\label{eq:info_nce}
	\mathcal{L}_{f}^{\mathbf{P}}=\,-\mathbb{E}_{\mathbf{p}_i\in \mathbf{P}_S}\left[\mathrm{log} \frac{f(\mathbf{p}_i,\mathbf{I}_{\mathbf{p}_i})}{f(\mathbf{p}_i,\mathbf{I}_{pr(i)})+\sum_{\mathbf{I}_{pr(i)}^{-}}f(\mathbf{p}_i,\mathbf{I}_{pr(i)}^{-})}\right],
\end{equation}
where $pr(i)$ is the index of pixel corresponding to point $\mathbf{p}_i$, $\mathbf{I}_{pr(i)}$ and $\mathbf{I}_{pr(i)}^{-}$ denote pixels that match and do not match point $\mathbf{p}_i$, respectively. These pairs are determined according to the distance between the projection of points and pixels on the image plane. We denote the positive and negative margins as $r_p$ and $r_n$.

Here, we adopt the log-bilinear model for $f(\cdot,\cdot)$ following ~\cite{oord2018representation}. 
To address the disparity in features between different modalities, we implement a learnable linear transformation (LTA) on the features from both modalities before computing the dot product. This approach is chosen over the traditional method of calculating feature similarities using cosine distance directly.
The formula is expressed as:
\begin{equation}\label{eq:func_log_bilinear}
        f(\mathbf{p}_i, \mathbf{I}_j) = e^{(\widetilde{\mathbf{F}}_i^{\mathbf{p}} \mathbf{W}_{f} \widetilde{\mathbf{F}}^{\mathbf{I}\top}_j)/\tau} = e^{(\widetilde{\mathbf{F}}_i^{\mathbf{p}} \mathbf{Q} \mathbf{D} \mathbf{Q}^{\top} \widetilde{\mathbf{F}}^{\mathbf{I}\top}_j)/\tau},
\end{equation}
where $\mathbf{W}_f$ represents a learnable linear transformation matrix, $\tau$ is the temperature factor, and $\mathbf{Q}\mathbf{D}\mathbf{Q}^{\top}$ is the decomposition result of $\mathbf{W}_f$, $\widetilde{\mathbf{F}}^{\mathbf{I}}$ and $\widetilde{\mathbf{F}}^{\mathbf{P}}$ denote normalized image and point features.
In the implementation, we ensure the symmetry of $\mathbf{W}_f$ such that it can be decomposed according to the properties of spectral decomposition. And computing similarities between features, i.e., $\widetilde{\mathbf{F}}_i^{\mathbf{p}} \mathbf{W}_{f} \widetilde{\mathbf{F}}^{\mathbf{I}\top}_j$, is facilitated to be bidirectional. 
Similarly, we can obtain the discriminative loss $\mathcal{L}_{f}^{\mathbf{I}}$ for the image. 
This method, termed learnable transformation alignment, enables the alignment of features across different modalities through a learnable transformation.

We illustrate the learned features using cosine distance and learnable transformation alignment, respectively, in Fig.~\ref{fig:feature_tsne}. We can observe that it is non-trivial to drive the matching 2D and 3D features to be similar with respect to cosine distance due to the inherent domain gap between point clouds and images. In contrast, when we use the learnable transformation alignment for computing similarities during training, the 2D and 3D features are closer and evenly distributed in the transformed unified representation space. We further analyze the effect of learnable transformation alignment in Section~\ref{sec:ablation_study}.

\subsection{Building Dense Correspondence}\label{sec:dense_corr}
\noindent\textbf{Overlapping Area Detection.} In our 2D-3D neural calibration framework, an essential component is overlapping area detection. 
Due to differences in the operational principles and the field of view of the camera and LiDAR sensors, the image and point cloud do not perfectly overlap.
This module is responsible for identifying regions where the point cloud and image data overlap~\cite{ren2022corri2p}. Accurate detection of these overlapping areas is crucial for effective feature alignment and calibration accuracy.
Specifically, we detect the overlapping areas for input images and point clouds based on their corresponding fused features $\mathbf{F}^\mathbf{I}$ and $\mathbf{F}^\mathbf{P}$~\cite{ren2022corri2p}. We denote the $\mathbf{S}^\mathbf{P}$ and $\mathbf{S}^\mathbf{I}$ as the predicted probabilities of points and pixels in overlapping areas, respectively.

The points and pixels with scores higher than pre-set thresholds are considered inside the overlapping areas.
 We denote the estimated points and pixels in the overlapping areas as $\mathbf{P}_o \in \mathbb{R}^{N_\mathbf{P}\times3}$ and $\mathbf{I}_o\in \mathbb{R}^{N_\mathbf{I}\times3}$, where ${N_\mathbf{P}}$ and ${N_\mathbf{I}}$ are the number of points and pixels in the overlapping area.

During the training phase, the model supervises predicted overlap scores through binary cross-entropy loss.
The loss function of overlapping area detection is defined as follows:
\begin{equation}
	\resizebox{0.48\textwidth}{!}{
		$\begin{aligned}
			\mathcal{L}{o} =& -\frac{1}{N} \sum_{i=1}^{N} \left[ \mathbb{I}_{o}(i) \log(\mathbf{S}^\mathbf{P}_{i}) + (1 - \mathbb{I}_{o}(i)) \log(1 - \mathbf{S}^\mathbf{P}_{i}) \right] \\
			&- \frac{1}{H'\times W'} \sum_{j=1}^{H'\times W'} \left[ \mathbb{I}_{o}(j) \log(\mathbf{S}^\mathbf{I}_{j}) + (1 - \mathbb{I}_{o}(j)) \log(1 - \mathbf{S}^\mathbf{I}_{j}) \right],
		\end{aligned}$
	}
\end{equation}
where \( \mathbb{I}_{o}() \) is an indicator function that equals 1 if the point $\mathbf{p}_i$ or pixel $\mathbf{I}_j$ is in the overlapping area and 0 otherwise.

\vspace{0.5em}
\noindent\textbf{Soft Matching.}
Previous work simply applies the hard assignment strategy, i.e., non-differentiable argmax operation, to build correspondences between the points and pixels, i.e., assigning points in the overlapped area to the pixels most similar to them~\cite{ren2022corri2p}. To make the overall framework end-to-end differentiable, here we predict the location of points in the image plane as a weighted sum of the target pixel coordinates:

\begin{equation}
	\widehat{\mathbf{C}}_{i}^{\mathbf{P}} = \sum_{\mathbf{I}_j \in \mathbf{I}_o} w_{ij} \cdot \mathbf{C}_{j}^{\mathbf{I}},
\end{equation}
where \( \widehat{\mathbf{C}}_{i}^{\mathbf{P}} \) represents the predicted coordinate of the point cloud point \( \mathbf{p}_i \) in the image plane, and \( \mathbf{C}_{j}^{\mathbf{I}} \) denotes the 2D coordinate of the \( j\)-th pixel. The weight \( w_{ij} \) signifies the feature similarity between the point \( \mathbf{p}_i \) and the pixel \( \mathbf{I}_j \), and is calculated based on the soft assignment principle:

\begin{equation}
	w_{ij} = \frac{\exp(\widetilde{\mathbf{F}}_i^{\mathbf{p}} \mathbf{W}_{f} \widetilde{\mathbf{F}}^{\mathbf{I}\top}_j)}{\sum_{\mathbf{I}_k \in \mathbf{I}_o} \exp(\widetilde{\mathbf{F}}_i^{\mathbf{p}} \mathbf{W}_{f} \widetilde{\mathbf{F}}^{\mathbf{I}\top}_k)}.
\end{equation}
This softmax-based weighting scheme ensures that the contribution of each pixel to the final predicted point location is proportional to its similarity with the point cloud point. Using the softmax function in this context allows for a differentiable and probabilistic approach to assigning correspondences, as opposed to the hard assignment methods used in previous works. 

\subsection{Estimation of Rigid Pose}\label{sec:estimate_pose}
The final module in our framework estimates the \revise{rigid pose} based on the soft matching results computed in Section~\ref{sec:dense_corr}. 
The neural calibration problem is defined as determining the \revise{rigid pose} $\{\mathbf{R}, \mathbf{t}\}$ that best maps points in overlapping areas $\mathbf{P}_o$ onto the estimated 2D coordinates in the image plane, i.e., 
\begin{equation}
	\widehat{\mathbf{R}}, \widehat{\mathbf{t}} = \mathop{\arg\min}_{\mathbf{R}, \mathbf{t}} \sum_{\mathbf{p}_i \in \mathbf{P}_o}\left[ \mathbf{K}(\mathbf{R}\mathbf{p}_i + \mathbf{t}) - \widehat{\mathbf{C}}_{i}^{\mathbf{P}} \right],
 \label{eq:problem}
\end{equation}
\revise{where $\mathbf{K}$ denotes the camera intrinsic matrix}.

In this paper, we solve Eq.~\eqref{eq:problem} using a differentiable EPnP algorithm~\cite{lepetit2009ep}.
The \revise{rigid pose} estimation is optimized by the loss function defined as:
\begin{equation}
	\mathcal{L}{p} = {\left \| \mathbf{R}_{gt}^{\top} \widehat{\mathbf{R}} - \mathbf{E} \right \|}_H + {\left \| \mathbf{t}_{gt} -\widehat{\mathbf{t}} \right \|}_H,
 \label{eq:loss_pose}
\end{equation}
where ${\left \| \cdot \right \|}_H $ is the Huber-loss, $\{\mathbf{R}_{gt}, \mathbf{t}_{gt}\}$ indicates the ground-truth pose, $\mathbf{E}$ is an identify matrix. 

\subsection{Training Pipeline}
Initially, each pair of an image and a point cloud is associated with \minor{the extrinsic parameters}, denoted as $\{\mathbf{R}_{raw}, \mathbf{t}_{raw}\}$.
Since the camera and LiDAR sensors are fixed in a relatively close position of the cars, the original pose aligning the camera coordinate to the point cloud coordinate is similar. To avoid the network overfitting to such scenarios, we simulate a range of sensor placements and orientations by applying random rotation $\mathbf{R}_r$ and translation $\mathbf{t}_r$ to the original point cloud during the pre-training process. The ground-truth pose is recalculated to accommodate these adjustments:
\begin{equation}\label{eq:pose_gt}
	\begin{aligned}
		\mathbf{t}_{gt}&=\mathbf{t}_{raw} - \mathbf{R}_{raw}\mathbf{t}_r, \\
		\mathbf{R}_{gt}&=\mathbf{R}_{raw}\mathbf{R}_r^{-1}.
	\end{aligned}
\end{equation}

Our network is trained end-to-end, with supervision from the ground truth poses $\{\mathbf{R}_{gt}, \mathbf{t}_{gt}\}$. The loss function is a weighted sum of three components: $\mathcal{L}=\lambda_f \mathcal{L}_f + \lambda_o \mathcal{L}_o + \lambda_p \mathcal{L}_p$. These components are weighted as follows: $\lambda_f=1.0$, $\lambda_o=0.5$, and $\lambda_p=0.2$. This training strategy ensures that our network learns to accurately estimate poses under varied conditions, enhancing the generalizability and effectiveness of learned representation.

\section{Experiments}\label{sec:experiments}
In this section, we begin with an overview of the datasets and evaluation metrics in Sec.\ref{sec:dataset}. Following this, Sec.\ref{sec:exp_semantic} to Sec.\ref{sec:exp_panoptic} detail the experimental setup and the fine-tuning results for three downstream tasks. Subsequent sections, Sec.\ref{sec:ablation_study} and Sec.\ref{sec:feature_matching_acc}, delve into comprehensive ablation studies and in-depth analyses to assess critical aspects of our framework. Finally, we explore the reasons why the pre-text task is effective in Sec~\ref{sec:discussions}.

\subsection{Dataset}\label{sec:dataset}
In this part, we briefly describe the datasets used for pre-training and fine-tuning downstream tasks.

\vspace{0.5em}
\noindent\textbf{SemanticKITTI (SK)} dataset contains RGB image and point cloud data pairs from KITTI scenes for the task of urban scene semantic segmentation~\cite{behley2019semantickitti}. The data were collected from sensor systems mounted on a car, comprising over 200,000 images and corresponding point clouds from 21 different scenes/sequences. \minor{The images and point clouds are synchronized, and their relative pose remains constant.} The original images have a resolution of 1241x376 pixels. Each point cloud contains approximately 40,000 3D points. Following common practice, the dataset is split into a training set using the 10 sequences and a validation set using the eighth sequence.

\setlength{\tabcolsep}{6pt}
\begin{table*}[b]
	\centering
	\caption{Results of semantic segmentation models fine-tuned on three distinct datasets. This comparison considers the quantity of annotated data, the dataset employed for pre-training, and variations in model architecture. Our analysis contrasts NCLR with a baseline that lacks pre-training and other contemporary self-supervised methods. \major{We report the mean Intersection over Union (mIoU) as a percentage metric and the performance gain relative to no pre-training.} The best results are highlighted in bold.}
	\label{table:semantic_seg_eval}
	\begin{tabular}{lll|ll|ll|ll|ll|ll}
		\toprule
		Dataset & Backbone   & Method  & \multicolumn{2}{c}{0.1\% ~} & \multicolumn{2}{c}{1\% ~} & \multicolumn{2}{c}{10\% ~} & \multicolumn{2}{c}{50\% ~} & \multicolumn{2}{c}{100\%}  \\ 
		\midrule
		\multirow{8}{*}{nuScenes~\cite{caesar2020nuscenes}}        & \multirow{5}{*}{MinkUNet~\cite{choy20194d}}    & No pre-training  & 21.6 & ~  & 35.0  & ~ & 57.3 & ~  & 69.0   & ~  & 71.2 &    \\
		&  & PointContrast~\cite{xie2020pointcontrast} & \textbf{27.1} & \textbf{+5.5} & 37.0 & +2.0 & 58.9 & +1.6 & 69.4 & +0.4 & 71.1 & -0.1 \\
		&  & DepthContrast~\cite{zhang2021depthcontrast} & 21.7 & +0.1 & 34.6 & -0.4 & 57.4 & +0.1 & 69.2 & +0.2 & 71.2 & 0 \\
		&  & ALSO~\cite{boulch2023ALSO}               & 26.2 & +4.6 & 37.4 & +2.4 & 59.0 & +1.7 & 69.8 & +0.8 & 71.8 & +0.6 \\
		&  & NCLR (Ours)       & 26.6 & +5.0 & \textbf{37.8} & \textbf{+2.8} & \textbf{59.5} & \textbf{+2.2} & \textbf{71.2} & \textbf{+2.2} & \textbf{72.7} & \textbf{+1.5} \\
		\cmidrule{2-13}
		& \multirow{3}{*}{SPVCNN~\cite{tang2020searching}}         & No     pre-training  & 22.2 & ~    & 34.4 & ~ & 57.1 & ~  & 69.0   & ~  & 70.7 &  \\
		&  & ALSO~\cite{boulch2023ALSO}      & 24.8 & +2.6 & 37.4 & +3.0 & 58.4 & +1.3 & 69.5 & +0.5 & 71.3 & +0.6  \\
		&  & NCLR (Ours)        & \textbf{25.8} & \textbf{+3.6}  & \textbf{37.8} & \textbf{+3.4} & \textbf{59.2} & \textbf{+2.1}  & \textbf{71.0} & \textbf{+2.0}  & \textbf{73.0} & \textbf{+2.3} \\ 
		\midrule
		\multirow{9}{*}{SemanticKITTI~\cite{behley2019semantickitti}} & \multirow{6}{*}{MinkUNet~\cite{choy20194d}} & No pre-training  & 30.0 & ~  & 46.2 & ~ & 57.6 & ~  & 61.8 & ~  & 62.7 &  \\
		&  & PointContrast~\cite{xie2020pointcontrast} & 32.4 & +2.4 & 47.9 & +1.7 & 59.7 & +2.1 & 62.7 & +0.9 & 63.4 & +0.7  \\
		&  & DepthContrast~\cite{zhang2021depthcontrast} & 32.5 & +2.5 & 49.0 & +2.8 & 60.3 & +2.7 & 62.9 & +1.1 & \textbf{63.9} & \textbf{+1.2}  \\
		&  & SegContrast~\cite{nunes2022segcontrast}   & 32.3 & +2.3 & 48.9 & +2.7 & 58.7 & +1.1 & 62.1 & +0.3 & 62.3 & -0.4  \\
		&  & ALSO~\cite{boulch2023ALSO}               & 35.0 & +5.0 & 50.0 & +3.8 & 60.5 & +2.9 & 63.4 & +1.6 & 63.6 & +0.9 \\
		&  & TARL~\cite{nunes2023TARL}               & 37.9 & +7.9 & 52.5 & +6.3 & 61.2 & +3.6 & 63.4 & +1.6 & 63.7 & +1.0 \\
		&  & NCLR (Ours)       & \textbf{39.2} & \textbf{+9.2} & \textbf{53.4} & \textbf{+7.2} & \textbf{61.4} & \textbf{+3.8} & \textbf{63.5} & \textbf{+1.7} & \textbf{63.9} & \textbf{+1.2} \\
		\cmidrule{2-13}
		& \multirow{3}{*}{SPVCNN~\cite{tang2020searching}}         & No pre-training  & 30.7 & ~    & 46.6 & ~ & 58.9 & ~  & 61.8 & ~  & 62.7 &  \\
		&  & ALSO~\cite{boulch2023ALSO}      & 35.0 & +4.3 & 49.1 & +2.5 & 60.6 & +1.7 & 63.6 & +1.8 & 63.8 & +1.1  \\
		&  & NCLR (Ours)        & \textbf{38.8} & \textbf{+8.1} & \textbf{52.8} & \textbf{+6.2} & \textbf{61.1} & \textbf{+2.2} & \textbf{64.0} & \textbf{+2.2} & \textbf{64.1} & \textbf{+1.4} \\ 
		\midrule
		\multirow{6}{*}{\makecell*[l]{SemanticPOSS~\cite{pan2020semanticposs}\\ (\major{pre-training on} \\ \major{SemanticKITTI})}}       & \multirow{6}{*}{MinkUNet~\cite{choy20194d}} & No pre-training  & 36.9 & ~    & 46.4 & ~ & 54.5 & ~  & 55.3 & ~  & 55.1 &  \\
		&  & PointContrast~\cite{xie2020pointcontrast} & 39.3 & +2.4 & 48.1 & +1.7 & 55.1 & +0.6 & 56.2 & +0.9 & 56.2 & +1.1  \\
		&  & DepthContrast~\cite{zhang2021depthcontrast} & 39.7 & +2.8 & 48.5 & +2.1 & 55.8 & +1.3 & 56.0 & +0.7 & 56.5 & +1.4  \\
		&  & SegContrast~\cite{nunes2022segcontrast}   & \textbf{41.7} & \textbf{+4.8} & 49.4 & +3.0 & 55.4 & +0.9 & 56.2 & +0.9 & 56.4 & +1.3  \\
		&  & ALSO~\cite{boulch2023ALSO}               & 40.7 & +3.8 & 49.6 & +3.2 & 55.8 & +1.3 & 56.4 & +1.1 & \textbf{56.7} & \textbf{+1.6}  \\
		&  & NCLR (Ours)       & \textbf{41.7} & \textbf{+4.8} & \textbf{49.8} & \textbf{+3.4} & \textbf{56.0} & \textbf{+1.5} & \textbf{56.6} & \textbf{+1.3} & \textbf{56.7} & \textbf{+1.6}  \\
		\bottomrule
	\end{tabular}
\end{table*}
\setlength{\tabcolsep}{1.5pt}

\vspace{0.5em}
\noindent\textbf{KITTI 3D Object Detection Dataset (KITTI3D)} is a prevailing collection of data specifically designed for 3D object detection in advancing autonomous driving technology~\cite{Geiger_KITTI}. The dataset is collected from various urban and suburban environments under different weather and lighting conditions. Each sample contains two modalities of 3D point clouds and RGB images.
The LiDAR sensor used in the KITTI dataset is a Velodyne HDL-64E LiDAR. The FOV of the camera in the KITTI dataset aligns closely with the range of the LiDAR sensor, ensuring comprehensive coverage of the surroundings of the vehicle.
Additionally, the dataset provides calibration information between the camera and the LiDAR sensor, which is essential for tasks that require the fusion of data from different sensors.
The dataset includes several types of objects commonly encountered in driving scenarios, such as cars, pedestrians, and cyclists. Each object in the dataset is annotated with a 3D bounding box, which provides precise information about the object's location, orientation, and dimensions in the 3D space.
Following common practice, we further split all training samples into a subset with 3712 samples for training and the remaining 3769 samples for validation. 
We evaluate performance using the Average Precision (AP) metric under IoU thresholds of 0.7, 0.5, and 0.5 for car, pedestrian, and cyclist categories, respectively. We compute APs with 40 sampling recall positions by default, instead of 11.

\vspace{0.5em}
\noindent\textbf{NuScenes Dataset} is comprised of driving footage captured in Boston and Singapore using a vehicle outfitted with a 32-beam LiDAR sensor and other sensors~\cite{caesar2020nuscenes}. Equipped with a full suite of sensors typical of autonomous vehicles, the dataset features a 32-beam LiDAR system, six cameras, and radar units, all providing a complete 360-degree coverage. 
The creators provide 850 total driving scene snippets, segmented into 700 training scene samples and 150 validation scene samples for the purposes of benchmarking models. 
Each of these scenes spans a duration of 20 seconds and is annotated at a frequency of 2 Hz.
The dataset provides detailed annotations for multiple object classes, such as vehicles, pedestrians, bicycles, and road barriers. Each object is annotated with a 3D bounding box, along with attribute information like visibility, activity, and pose. It has also been extended to include capabilities for semantic segmentation and panoptic segmentation, known as nuScenes-seg. 

In an effort to enable fair evaluations against previously published models, we used 600 of the 700 training scenes to pretrain our model, reserving the leftover 100 training scenes to calibrate hyperparameters. After this tuning period, model performance is measured using the full 150 validation scenes provided in the dataset. This validation set therefore functions as our test set. 
It is useful to note that the nuScenes data includes LiDAR sweeps that have not been manually annotated - the sensors record at 20 Hz, while human annotations are only provided for every 10th LiDAR sweep. These unannotated scans could be utilized in a self-supervised manner to pretrain models. However, to stay consistent with prior work, we only leveraged the annotated scans during our pre-training phase, while ignoring the annotations themselves. After pretraining, we fine-tuned our model on various subsets of the 700 scenes in the training set. More details of this precise evaluation protocol can be found in these published benchmark papers ~\cite{caesar2020nuscenes}. In our experiments, we followed this same protocol to allow for standardized comparisons.


\vspace{0.5em}
\noindent\textbf{SemanticPOSS Dataset} is a valuable asset for 3D semantic segmentation studies~\cite{pan2020semanticposs}, consisting of 2988 diverse and complex LiDAR scans featuring numerous dynamic instances. Collected at Peking University, it conforms to the SemanticKITTI data format. This dataset is especially relevant for autonomous driving research, encompassing 14 categories like people, riders, and cars. In line with the official evaluation guidelines, the third sequence is designated as the validation set, while the remaining sequences form the training set.

\begin{figure*}[t]
	\centering
	\includegraphics[width=\textwidth]{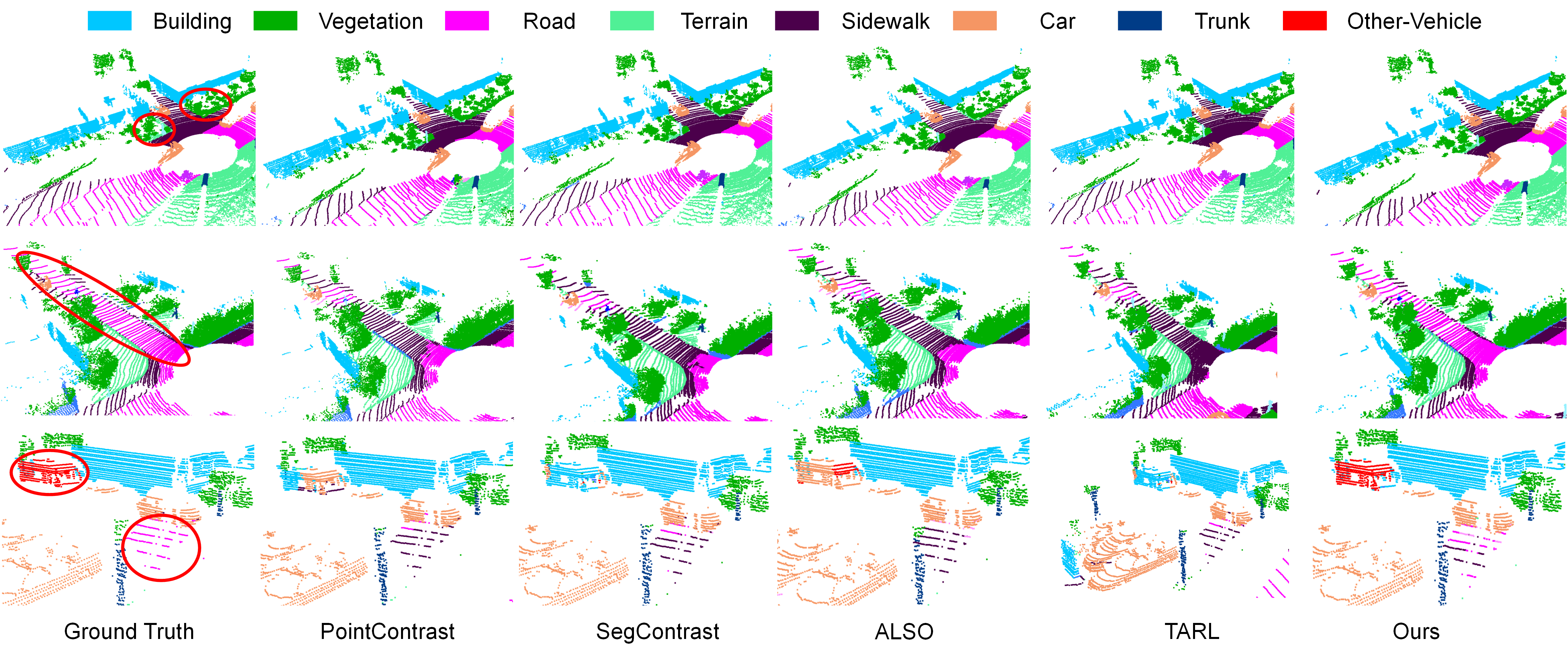} 
	\caption{
		\major{Qualitative analysis of semantic segmentation on SemanticKITTI: fine-tuning performance with 1\% of annotated training data. We highlight some areas that are easily misclassified with red circles.}
	}
	\label{fig:semantic_seg_vis}
\end{figure*}

\setlength{\tabcolsep}{5pt}
\begin{table}[t]
	\centering
	\caption{Comparison of various pre-training techniques for semantic segmentation tasks using either finetuning or linear probing. We report the mean Intersection over Union (mIoU) on the nuScenes validation set across different annotation levels.
	}
	\label{table:comparison_multimodal_contrastrive}
	\begin{tabular}{c|c|cc|c} 
		\toprule
		\multirow{3}{*}{Initialization} & \multirow{3}{*}{Reference} & \multicolumn{2}{c|}{nuScenes} & SemanticKITTI  \\
		&    & Lin. Prob & Finetune          & Finetune       \\
		&  & 100\%    & 1\%  & 1\%   \\ 
		\hline
		Random                           & -          & 8.10    & 30.30  & 39.50   \\
		PPKT~\cite{li2022simipu}         & Arxiv'21   & 35.90   & 37.80  & 44.00     \\
		SLidR~\cite{sautier2022image}    & CVPR'22    & 38.80   & 38.30  & 44.60   \\
		ST-SLidR~\cite{mahmoud2023self}  & CVPR'23    & 40.48  & 40.75 & 44.72  \\ 
		\hline
		Ours                             & -          & \textbf{42.81}  & \textbf{42.29} & \textbf{45.84}  \\
		\bottomrule
	\end{tabular}
\end{table}
\setlength{\tabcolsep}{7pt}

\subsection{Semantic Segmentation}\label{sec:exp_semantic}

\noindent\textbf{Network Architectures.} 
To assess the versatility of our method across various architectures and to ensure an equitable comparison with prior studies, we conducted experiments using multiple backbone networks. Our tests included two versions of MinkUNet~\cite{choy20194d}, along with ResUNet18 applied to SemanticKITTI, ResUNet34 for nuScenes, and the SPVCNN~\cite{tang2020searching}.
For the image branch, we utilized the ResNet backbone with the FPN structure to extract features. During the fine-tuning process for semantic segmentation, the image branch was omitted, and we substituted the final layer of the point cloud backbone. This replacement involves introducing a new fully-connected layer, designed with the channel matching the number of segmentation classes. This modification facilitates per-point predictions specifically adapted for the segmentation task at hand.

\vspace{0.5em}
\noindent\textbf{Pre-training Protocol.} 
We pre-trained the backbones for 50 epochs with a total batch size of 32 distributed across 8 GPUs for SemanticKITTI, and a batch size of 16 distributed on 4 GPUs for the nuScenes dataset.
We trained the network with the AdamW optimizer~\cite{loshchilov2017decoupled}, a learning rate of 0.001, a weight decay of 0.001, a temperature hyperparameter $\tau$ of 0.07 in Eq.~\eqref{eq:func_log_bilinear}, and a cosine annealing scheduler.
\major{Some models were pre-trained on the SemanticKITTI dataset and then evaluated on both SemanticKITTI and SemanticPOSS, while others were pre-trained and evaluated on the nuScenes dataset.}
During the pre-training process, all points, including those labeled as ``ignore" are taken as input, and those not labeled points were filtered out for downstream fine-tuning.
The input range on the $x$-$y$ plane is [-51.2m, 51.2m] and [-3m, 1m] on the $z$-axis. Point clouds were voxelized with a grid size of 0.05m and 0.1m for SemanticKITTI and nuScenes, respectively.
During training, the image size was resized to $160\times 512$ and $160\times 320$ for SemanticKITTI and nuScenes, respectively. The point clouds underwent random transformations, including rotation around the $z$-axis and 2D translation in the $x$-$y$ plane, where the rotation angle and translation of each direction were uniformly drawn from [-$\pi$, $\pi$] and [-15m, 15m], respectively.

\vspace{0.5em}
\noindent\textbf{Downstream Fine-tuning Protocol.} Following the pre-training phase, we fine-tuned the neural network backbones specifically for the task of semantic segmentation. This fine-tuning was conducted on selected data subsets from the SemanticKITTI and nuScenes datasets. We adhered to the fine-tuning hyperparameters and dataset splits as detailed in ALSO~\cite{boulch2023ALSO}. In the case of nuScenes, we implemented a batch size of 8, and for SemanticKITTI, we opted for a smaller batch size of 2. The final per-point scoring was determined by assigning each point the prediction of the voxel in which it is located. It is worth noting that the epoch count varies in relation to the percentage of training data used: for 0.1\% of the data, 1000 epochs were used; for 1\%, it's 500 epochs; 100 epochs were applied to 10\% of the data; 50 epochs for 50\%; and 30 epochs when training with the full dataset.

\vspace{0.5em}
\noindent\textbf{Quantitative Evaluation Results.} 
Table~\ref{table:semantic_seg_eval} presents the experimental results experiments on fine-tuning semantic segmentation models using three distinct datasets: nuScenes, SemanticKITTI, and SemanticPOSS.
We observe that NCLR consistently outperforms other methods across different data volumes, showing particularly strong gains over the baseline training from scratch. For instance, with 0.1\% of the data, NCLR achieves a 5.0\% and 3.6\% improvement in mIoU over the baseline for MinkUNet and SPVCNN, respectively. Similarly, NCLR demonstrates substantial improvements on the SemanticKITTI dataset, especially with limited training data. For MinkUNet, it shows a 9.2\% increase in mIoU with just 0.1\% of the data compared to the baseline. Similar trends are observed with SPVCNN. 
In the SemanticPOSS dataset, NCLR exhibits superior performance compared to traditional non-pre-trained models and other self-supervised approaches. The improvement is consistent across all data levels, demonstrating the generalization and effectiveness of NCLR in learning useful representations for 3D semantic segmentation tasks.
\major{The pose ground truth used is derived from the known sensor setup and applied data augmentation, rather than from additional costly manual annotation, so the comparisons with state-of-the-art methods are fair.}

\vspace{0.5em}
\noindent\textbf{Qualitative Results.} Fig.~\ref{fig:semantic_seg_vis} illustrates qualitative results derived from various methods. 
Our self-supervised pre-training method produces more accurate segmentation results compared to other self-supervised learning approaches. The segmentation results from our method contain more precise boundaries and finer details. Specifically, it achieves better performance at distinguishing between visually similar object classes like the road and sidewalk.

\vspace{0.5em}
\noindent\textbf{\revise{Additional Comparison with Cross-modal Contrastive Methods.}} In Table~\ref{table:semantic_seg_eval}, we evaluate the semantic segmentation based on the settings used by ALSO~\cite{boulch2023ALSO}. However, those settings differ from the ones adopted in contrastive methods such as~\cite{li2022simipu,sautier2022image,mahmoud2023self}. To align with the settings adopted in these contrastive methods, we have conducted additional experiments based on their experimental protocols. \major{We used momentum SGD for optimization with an initial learning rate of 0.5, momentum of 0.9, and weight decay of 1e-4. The learning rate was adjusted using a cosine annealing scheduler over 50 epochs. For fine-tuning, we followed previous works and fine-tuned the network for 100 epochs with batch sizes of 10 and 16 on the SemanticKITTI and nuScenes LiDAR segmentation datasets, respectively. The initial learning rates for the backbone and the segmentation head were set to 0.05 and 2.0, respectively.}
As presented in Table~\ref{table:comparison_multimodal_contrastrive}, our approach also outperforms these baselines across the nuScenes and SemanticKITTI datasets. Specifically, our method shows significant improvements in both linear probing and fine-tuning settings. For linear probing with 100\% of annotations on nuScenes, our approach achieves 42.81\% mIoU, surpassing the previous state-of-the-art ST-SLidR (40.48\%). When fine-tuning with only 1\% of annotations, our method achieves a mIoU of 42.29\%, outperforming ST-SLidR (40.75\%). Furthermore, our method demonstrates strong performance on the SemanticKITTI dataset, reinforcing its generalization capability across different datasets and tasks.

\setlength{\tabcolsep}{5pt}
\begin{table*}[htbp]
	\centering
	\caption{
		Comparisons between our method and other self-supervised learning methods fine-tuned on the KITTI3D dataset.
		We report the AP evaluated with 40 recall positions on the val set of the KITTI3D dataset. Note ProposalContrast~\cite{yin2022proposalcontrast} is specifically tailored for 3D object detection and pre-trained with the large-scale Waymo dataset. The best results are highlighted in bold.}
	\label{table:3dod_kitti3d}
	\begin{tabular}{c|l|ccc|ccc|ccc|c} 
		\toprule
		\multirow{2}{*}{Detector} & \multirow{2}{*}{\makecell{Pre-training \\Schedule}} & \multicolumn{3}{c|}{Car}        & \multicolumn{3}{c|}{Pedestrian} & \multicolumn{3}{c|}{Cyclist}   & \multirow{2}{*}{Avg}  \\
		& ~ & Easy  & Moderate & Hard & Easy  & Moderate & Hard & Easy  & Moderate & Hard &                       \\ 
		\midrule
		\multirow{3}{*}{SECOND}   & Scratch   & 90.20    & 81.50 & 78.61 & 53.89   & 48.82    & 44.56 & 82.59   & 65.72    & 62.99 & 67.65 \\
		& ALSO~\cite{boulch2023ALSO}       & 90.20    & 81.53    & 78.83 & 57.30    & 53.21    & 48.32 & 82.92   & 69.12    & 64.57 & 69.56 \\
		& NCLR (Ours)   & \textbf{90.23}   & \textbf{81.99}    & \textbf{79.05} & \textbf{59.20}   & \textbf{54.75}    & \textbf{49.32} & \textbf{83.64}   & \textbf{70.16 }   & \textbf{65.13} & \textbf{70.38} \\ 
		\midrule
		\multirow{6}{*}{PV-RCNN}  & Scratch   & 91.74   & 84.60 & 82.29 & 65.51   & 57.49    & 52.71 & 91.37   & 71.51    & 66.98 & 73.80 \\
		& ALSO~\cite{boulch2023ALSO}       & 92.15   & 84.85    & \textbf{82.59} & 65.63   & 57.83    & 53.14 & 91.81   & 73.85    & 69.71 & 74.62 \\
		& STRL~\cite{huang2021STRL}       & -   & 84.70 & -     & -   & 57.80 & -     & -   & 71.88    & -     &  -     \\
		& PointContrast~\cite{xie2020pointcontrast}  & 91.40    & 84.18    & 82.25 & 65.73   & 57.74    & 52.46 & 91.47   & 72.72    & 67.95 & 73.99 \\
		& ProposalContrast~\cite{yin2022proposalcontrast} & \textbf{92.45}   & 84.72    & 82.47 & \textbf{68.43}   & 60.36    & 55.01 & \textbf{92.77}   & 73.69    & 69.51 & 75.49 \\
		& NCLR (Ours)   & 92.43 & \textbf{84.86}  & 82.58   & 67.89 & \textbf{60.48}   & \textbf{55.47}   & 92.45 & \textbf{74.05}  & \textbf{70.29}   & \textbf{75.61} \\
		\bottomrule
	\end{tabular}
\end{table*}

\setlength{\tabcolsep}{4pt}
\begin{table}[htbp]
	\centering
	\caption{
		The results (mAP) for models fine-tuned with varying quantities of annotated data. These results specifically reflect the performance in 3D object detection under moderate difficulty conditions on the validation set of the KITTI3D dataset.
	}
	\label{table:label_efficient_3dod}
	\begin{tabular}{l|l|l|lll|l}
		\toprule 
		Ratio & Detector & \makecell{Pre-training \\Schedule} & Car & Ped. & Cyc. & mAP (\%)  \\
		\midrule
		\multirow{4}{*}{10\%} & \multirow{2}{*}{SECOND}   & Scratch      & 75.02 &  40.53 & 46.09 & 53.88    \\
		&          & NCLR (Ours) & 76.57 &  42.69 & 49.69 & \bf56.31{\scriptsize (+2.43)}    \\
		\cmidrule{2-7}
		& \multirow{2}{*}{PV-RCNN}  & Scratch  & 79.65 &  51.42 & 61.06 & 64.04  \\
		&          & NCLR (Ours) & 82.32 &  55.95 & 62.52 & \bf66.93{\scriptsize (+2.89)}   \\
		\midrule
		\multirow{4}{*}{20\%} & \multirow{2}{*}{SECOND}   & Scratch   & 78.12 &  42.35 & 60.97 & 60.48  \\
		&          & NCLR (Ours) & 79.17 &  44.62 & 64.78 & \bf62.85{\scriptsize (+2.37)} \\
		\cmidrule{2-7}
		& \multirow{2}{*}{PV-RCNN}  & Scratch    & 82.37 &  53.70 & 67.31 & 67.79\\
		&          & NCLR (Ours) & 82.56 &  57.29 & 69.92 & \bf69.92{\scriptsize (+2.13)}  \\
		\midrule
		\multirow{4}{*}{50\%} & \multirow{2}{*}{SECOND}   & Scratch    & 80.84 &  46.55 & 64.02 & 63.81    \\
		&          & NCLR (Ours) & 81.80 &  49.38 & 66.27 & \bf65.81{\scriptsize (+2.00)}  \\
		\cmidrule{2-7}
		& \multirow{2}{*}{PV-RCNN}  & Scratch & 82.45 &  56.87 & 70.36 & 69.89  \\
		&          & NCLR (Ours)   & 82.95 &  58.91 & 73.67 & \bf71.84{\scriptsize (+1.95)}  \\
		\bottomrule
	\end{tabular}
\end{table}

\renewcommand{\arraystretch}{1.25}
\setlength{\tabcolsep}{8pt}
\begin{table*}[htbp]
	\centering
	\caption{\revise{Comparisons between our method and other pre-training methods by finetuning on 5\% of the Waymo 3D detection dataset. We report the mAP and mAPH metrics at LEVEL\_2 on the validation set. Random initialization denotes training from scratch. The best results are highlighted in bold.}}
	\label{table:comparison_waymo}
	\begin{tabular}{c|cc|cccccc} 
		\toprule
		\multirow{2}{*}{Initialization} & \multicolumn{2}{c|}{Overall} & \multicolumn{2}{c}{Vehicle} & \multicolumn{2}{c}{Pedestrian} & \multicolumn{2}{c}{Cyclist}  \\
		& mAP   & mAPH  & mAP   & mAPH & mAP  & mAPH  & mAP   & mAPH  \\ 
		\hline
		Random                                           & 43.68 & 40.29 & 54.05 & 53.50 & 53.45 & 44.76   & 23.54 & 22.61 \\
		PointContrast~\cite{xie2020pointcontrast}        & 45.32 & 41.30 & 52.12 & 51.61 & 53.68 & 43.22   & 30.16 & 29.09 \\
		ProposalContrast~\cite{yin2022proposalcontrast}  & 46.62 & 42.58 & 52.67 & 52.19 & 54.31 & 43.82   & 32.87 & 31.72 \\
		MV-JAR~\cite{xu2023mv}                           & 50.52 & 46.68 & 56.47 & 56.01 & 57.65 & 47.69   & 37.44 & 36.33 \\
		PonderV2~\cite{zhu2023ponderv2}                  & 50.87 & 46.81 & 56.54 & 55.49 & 58.09 & 48.41   & 37.98 & 36.52 \\
            \hline
		Ours                                             & \textbf{51.86} & \textbf{48.32} & \textbf{57.13} & \textbf{55.93} & \textbf{59.32} & \textbf{50.15}   & \textbf{39.14} & \textbf{38.89}  \\
		\bottomrule
	\end{tabular}
\end{table*}
\setlength{\tabcolsep}{7pt}
\renewcommand{\arraystretch}{1.0}

\subsection{3D Object Detection}\label{sec:exp_3dod}
\noindent\textbf{Network Architectures.} 
In the downstream 3D object detection task, we explored two prevalent network architectures: the SECOND~\cite{yan2018second} and PV-RCNN~\cite{shi2020pv} object detectors. Both architectures are built upon a shared foundational design, which includes a 3D sparse encoder. This encoder processes the input voxels through 3D sparse convolutional operations. In addition, they incorporate a bird's-eye-view encoder (termed the 2D-backbone), which is activated post-BEV projection. The key difference between them lies in their respective detection heads. SECOND employs a region proposal network (RPN) directly on the 2D backbone, whereas PV-RCNN enhances the RPN predictions through point-level refinement. This refinement leads to more accurately defined bounding boxes and improved confidence estimations.

\vspace{0.5em}
\noindent\textbf{Pre-training Protocol.} 
In alignment with the requirements of downstream detection tasks, we processed the point clouds through a voxelization step. This involves setting the grid size to [0.05m, 0.05m, 0.1m] along the $x$, $y$, and $z$ axes, respectively. The maximum number of non-empty input voxels was limited to 60,000. The raw image was resized to (512, 160) as input.
We utilized the default AdamW optimizer. The settings of the optimizer include a peak learning rate of 0.001 and a weight decay factor of 0.001. Cosine learning rate schedule~\cite{loshchilov2016sgdr} was adopted. 
We pre-trained the backbone for 50 epochs on the SemanticKITTI dataset with a batch size of 8 on a single GPU.
Regarding the configuration of the VoxelNet backbone~\cite{zhou2018voxelnet}, we ensured consistency by employing parameters identical to those used in the corresponding 3D object detectors.

\vspace{0.5em}
\noindent\textbf{Downstream Fine-tuning Protocol.} In the subsequent stage, the detection module from either SECOND or PVRCNN was integrated with the pre-trained neural network, and the combined detector was further fine-tuned on the KITTI3D dataset. This process utilized the OpenPCDet framework\footnote{https://github.com/open-mmlab/OpenPCDet}, specifically its implementation of these detectors, along with the standard training parameters provided by OpenPCDet. Consistent with the methodology outlined in a prior study~\cite{boulch2023ALSO}, this fine-tuning process was repeated three times independently, and the highest mean Average Precision (mAP) achieved on KITTI3D's validation set was recorded and presented.

\setlength{\tabcolsep}{7pt}
\renewcommand{\arraystretch}{1.25}
\begin{table*}[t]
	\centering
	\caption{
            The performance metrics (PQ and IoU) for panoptic segmentation on SemanticKITTI showcase the effectiveness of the pre-trained models when fine-tuning with various proportions of annotated data. The best results are highlighted in bold.
            }
	\label{table:panoptic_seg_eval}
	\begin{tabular}{lcc|cc|cc|cc|cc} 
		\toprule
		\multirow{2}{*}{\makecell{Pre-training \\Schedule}}            & \multicolumn{2}{c}{0.10\% ~}                 & \multicolumn{2}{c}{~ 1\%}                    & \multicolumn{2}{c}{~ 10\%}                   & \multicolumn{2}{c}{50\% ~}                   & \multicolumn{2}{c}{100\% ~}                  \\ 
		\cline{2-11}
		& PQ    & IoU    & PQ    & IoU    & PQ    & IoU    & PQ    & IoU    & PQ    & IoU    \\ 
		\hline
		From Scratch            & 4.76  & 11.13  & 22.72 & 30.84  & 47.20 & 53.53  & 55.32 & 61.94  & 55.40 & 59.75   \\
		PointContrast~\cite{xie2020pointcontrast} & 5.86  & 11.51  & 27.37 & 32.49  & 47.57 & 54.63  & 54.21 & 59.48  & 55.85 & 61.49   \\
		DepthContrast~\cite{zhang2021depthcontrast} & 7.65  & 13.56  & 27.31 & 32.30  & 46.85 & 51.27  & 54.55 & 59.60  & 56.15 & 60.81   \\
		SegContrast~\cite{nunes2022segcontrast}   & 7.58  & 14.46  & 26.14 & 32.85  & 47.02 & 53.47  & 55.38 & 60.04  & 56.73 & 61.96   \\
		TARL~\cite{nunes2023TARL}               & 10.26 & 17.01  & 29.24 & 34.71  & 51.27 & 57.59  & 56.10 & 62.36  & 56.57 & 62.05   \\ 
		\hline
		NCLR (Ours)        & \textbf{12.79} & \textbf{19.14} & \textbf{30.69} & \textbf{36.24} & \textbf{52.79} & \textbf{58.78} & \textbf{56.78} & \textbf{62.54} & \textbf{56.87} & \textbf{62.69}  \\
		\bottomrule
	\end{tabular}
\end{table*}

\vspace{0.5em}
\noindent\textbf{Quantitative Evaluation Results.} 
We evaluate the transferability of our pre-trained model by first pre-training on the SemanticKITTI dataset and subsequently fine-tuning on the KITTI3D dataset. We report the experimental results in Table~\ref{table:3dod_kitti3d}.
It is evident that our approach consistently surpasses the baseline established by training from scratch, achieving an enhancement for both SECOND and PV-RCNN. 
With our pre-training method, the performance improves, particularly in the Pedestrian and Cyclist categories, reaching an average mAP of 70.38\%.
Our method performs comparably to state-of-the-art models like ProposalContrast~\cite{yin2022proposalcontrast}, which is specifically tailored for detection tasks and pre-trained with the large-scale Waymo dataset.

One of the primary benefits of self-supervised learning lies in its ability to improve the performance of downstream tasks when the annotation source is limited. 
To further assess this aspect, we evaluated our method in label-efficient 3D object detection. We consider a model trained from a state of random initialization as our standard for comparison. The superiority of our pre-trained model compared to this standard is detailed in Table~\ref{table:label_efficient_3dod}. Essentially, our pre-trained model consistently boosts the performance in detection tasks using both SECOND and PV-RCNN architectures, particularly notable under conditions of limited labeled data – showing an improvement of 2.43\% and 2.89\% in mAP with just 1\% labeled data for SECOND and PV-RCNN, respectively. In addition, our model surpasses the baseline performances across all variations of available annotated data quantities.

\revise{We compare NCLR with MAE-based MV-JAR~\cite{xu2023mv} and rendering-based PonderV2~\cite{zhu2023ponderv2} on the Waymo 3D detection benchmark. Specifically, we employ the official implementation of MV-JAR~\cite{xu2023mv} and SST~\cite{fan2022embracing} for pre-training and downstream evaluation, respectively. As shown in Table~\ref{table:comparison_waymo}, our method consistently outperforms MAE-based MV-JAR~\cite{xu2023mv} and PonderV2~\cite{zhu2023ponderv2} across most evaluation metrics on the Waymo 3D detection dataset when fine-tuning with 5\% of the available annotations. Specifically, our method achieves the best overall results, i.e., 51.86\% mAP and 48.32\% mAPH. It represents improvements of +0.99\% and +1.51\%, respectively, compared to the rendering-based baseline, PonderV2~\cite{zhu2023ponderv2}.}

\begin{table}[t]
	\centering
	\caption{Experimental results from models pre-trained through self-supervised and supervised training on SemanticKITTI, subsequently fine-tuned for panoptic segmentation tasks using both the mini and complete training sets of nuScenes. We report the PQ and IoU metrics on the full validation set of nuScenes.
	}
	\label{table:panoptic_seg_transfer}
	\begin{tabular}{l|cc|cc} 
		\toprule[1.2pt]
		\multirow{2}{*}{\makecell{Pre-training \\Schedule}}    & \multicolumn{2}{c|}{Mini} & \multicolumn{2}{c}{Full}  \\
		~                         & PQ    & IoU              & PQ    & IoU                 \\ 
		\midrule
		From Scratch              & 23.78 & 23.96            & 52.98 & 58.17               \\
		Supervised pre-training   & 24.77 & 23.6             & 53.19 & 58.05               \\
		\midrule
		PointContrast~\cite{xie2020pointcontrast}        & 26.58 & 25.46            & 51.06 & 56.39               \\
		DepthContrast~\cite{zhang2021depthcontrast}         & 28.66 & 27.3             & 51.51 & 57.06               \\
		SegContrast~\cite{nunes2022segcontrast}          & 28.84 & 26.79            & 52.31 & 57.24               \\
		TARL~\cite{nunes2023TARL}                      & 32.22 & 30.73            & 53.26 & 59.14               \\ 
		\midrule
		NCLR (Ours)     & \textbf{33.37} & \textbf{31.62} & \textbf{53.97} & \textbf{59.76}  \\
		\bottomrule[1.2pt]
	\end{tabular}
\end{table}

\subsection{Panoptic Segmentation}\label{sec:exp_panoptic}
To assess the instance-level features learned by our method, we further fine-tuned our pre-trained models for the task of panoptic segmentation.

\vspace{0.5em}
\noindent\textbf{Network Architectures.} In this part, we chose MINKUnet-34~\cite{choy20194d} as the 3D point-wise feature backbone. For the fine-tuning process, semantic and instance heads are integrated with the pre-trained 3D backbone, followed by clustering post-processing to identify the instances~\cite{nunes2023TARL}. 
This clustering stage leverages semantic predictions to exclude background points, focusing solely on foreground instances such as vehicles, pedestrians, and cyclists. Once the irrelevant points are removed, the remaining points undergo a clustering process to discern distinct instances based on the features from the instance head. For the clustering phase, we opt for the mean shift algorithm ~\cite{comaniciu2002mean} with a set bandwidth of 1.2 and a minimum cluster size requirement of 50 points.
 The other architectures and settings are the same as Section~\ref{sec:exp_semantic} during pre-training. 

\vspace{0.5em}
\noindent\textbf{Protocol for Fine-tuning on Downstream Tasks.} Our methodology employed the AdamW optimizer, utilizing a max learning rate set at 0.2. In the context of panoptic segmentation applied to the SemanticKITTI dataset, the fine-tuning of our model was conducted across various subsets of the annotated training data, specifically 0.1\%, 1\%, 10\%, 50\%, and the full 100\% subset. For the nuScenes dataset, the fine-tuning process used both the complete training set and the mini training subset that is provided. The performance of the fine-tuned model was then assessed on the complete validation set of all datasets. We fine-tuned the model for fixed 50 epochs.

\vspace{0.5em}
\noindent\textbf{Results of Fine-tuning.} We pre-trained the backbone on the SemanticKITTI dataset and fine-tuned it on different percentage subsets of the SemanticKITTI dataset. As shown in Table~\ref{table:panoptic_seg_eval}, our method is consistently better than previous self-supervised pre-training approaches.
When the segmentation model is trained with fewer labels, we can observe an obvious improvement compared to the baseline of scratch training.

\vspace{0.5em}
\noindent\textbf{Generalization of Learned Features.} Our study also investigated the adaptability of learned features focusing on panoptic segmentation tasks. We initially pre-trained the 3D backbone on the SemantiKITTI dataset and subsequently fine-tuned it on both the full and mini training sets of nuScenes.
As presented in Table~\ref{table:panoptic_seg_transfer}, while all approaches enhance performance on the nuScenes dataset, the features generated through our technique demonstrate greater adaptability when employed in a different domain dataset. Furthermore, in comparison to supervised pre-training on SemanticKITTI, our approach exhibits improved results. This finding underscores the enhanced effectiveness and potential of our method in scenarios of transfer learning, surpassing the conventional supervised pre-training methods.

In general, the experiments conducted further confirm that our approach effectively extracts semantic nuances and identifies instance-level details, achieving higher scores in IoU and PQ metrics than former methodologies.

\begin{figure*}[b]
	\centering
	\includegraphics[width=\textwidth]{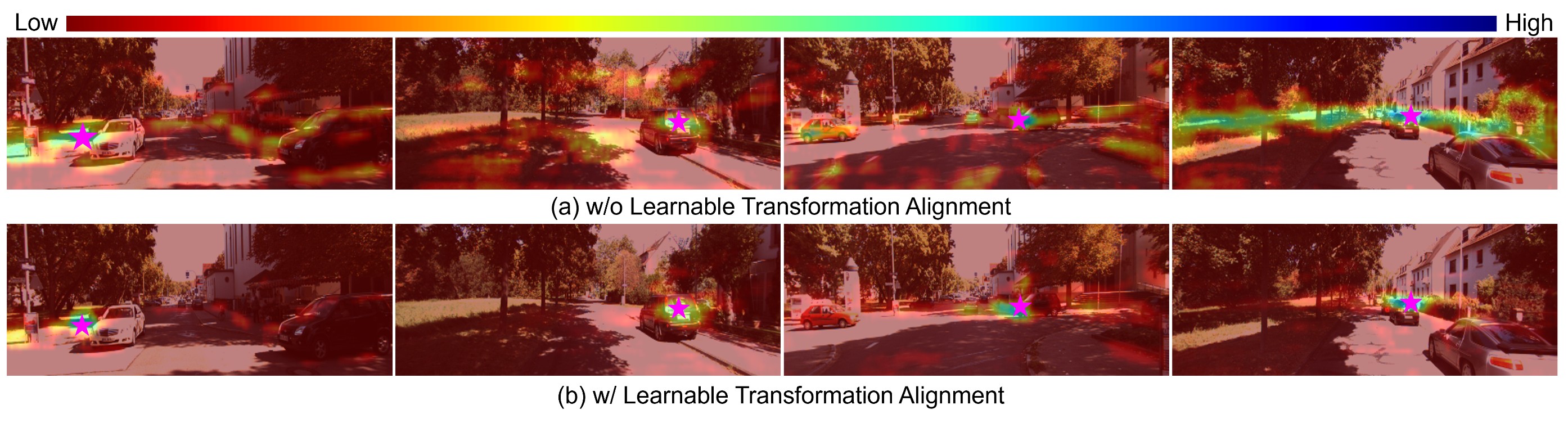} 
        \vspace{-0.5cm}
	\caption{
		\major{Visual comparison of the similarities between image and point cloud features learned by the model w/o and w/ learnable transformation alignment. For a selected 3D query point (indicated by a purple pentagram), we calculate its similarities with image features, forming a 2D similarity map. These similarity maps are illustrated using examples from the SemanticKITTI validation set.}
	}
	\label{fig:sk_3dquery_similarity_vis}
\end{figure*}

\begin{figure*}[t]
	\centering
	\includegraphics[width=\textwidth]{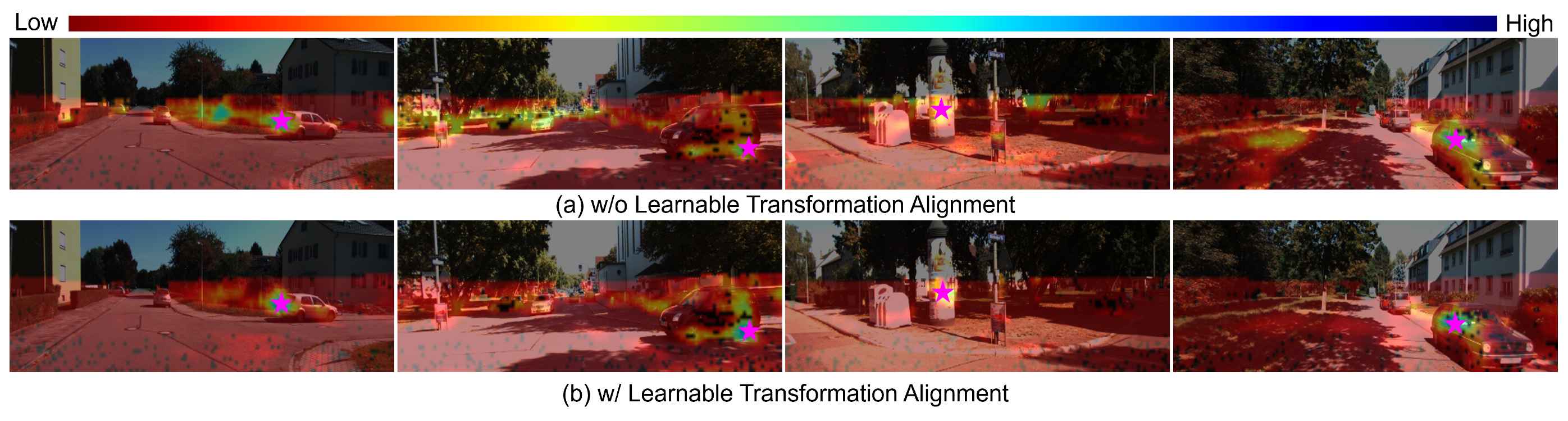} 
        \vspace{-0.7cm}
	\caption{
		\major{Visual comparison of the similarities between image and point cloud features learned by the model w/o and w/ learnable transformation alignment. For a selected 2D query pixel (marked with a purple pentagram), we calculate its similarities with point features and project these similarities onto the image plane, forming a similarity map. These similarity maps are illustrated using examples from the SemanticKITTI validation set.}
	}
	\label{fig:sk_2dquery_similarity_vis}
        \vspace{-0.2cm}
\end{figure*}

\subsection{Ablation Studies}\label{sec:ablation_study}

In our ablation studies on the nuScenes dataset, we adhered to the evaluation protocol from ALSO~\cite{boulch2023ALSO}. This involves splitting the nuScenes training set into two parts: ablation-train and ablation-val. Fine-tuning was performed with just 1\% of annotations from the ablation-train set. This ensures parameter tuning does not involve the validation set, reserved for comparing against other methods. We limited the training period for these studies to 100 epochs.

\setlength{\tabcolsep}{4pt}
\begin{table}[t]
	\centering
	\caption{Ablation study on nuScenes semantic segmentation. End-to-end pose estimation indicates using soft-matching, differentiable EPnP solver, and loss of pose estimation.}
	\label{table:ablation_study}
	\begin{tabular}{l|cccc}
		\toprule[1.2pt]
		Baseline                         & \checkmark & \checkmark & \checkmark & \checkmark  \\
		Learnable Transformation Alignment & $\times$ & \checkmark & \checkmark & \checkmark  \\
		Overlapping Area Detection       & $\times$ & $\times$ & \checkmark & \checkmark  \\
		End-to-End Pose Estimation      & $\times$ & $\times$ & $\times$ & \checkmark \\
		\midrule
		mIoU (\%)                  & 35.90 & 36.52 & 37.13 & \bf37.92 \\
		\bottomrule[1.2pt]
	\end{tabular}
\end{table}

\vspace{0.5em}
\noindent\textbf{Effect of Key Components.}
We analyze the impact of each component incorporated into our method in Table~\ref{table:ablation_study}. We can observe that our proposed key components benefit the pre-training framework in learning useful representations and yield better performance in downstream tasks. 
For example, the introduction of learnable transformation alignment increases the mIoU from 35.90\% to 36.52\%. This indicates that allowing the model to adaptively transform features enhances its ability to capture the inherently non-linear and complex relationship between point cloud data and RGB images.
Besides, the supervision of advanced tasks, including overlapping area detection and pose estimation also prompts the network to fully understand the two modalities and learn useful representation.
Then we analyze the effect of learnable transformation alignment in detail.

\setlength{\tabcolsep}{8pt}
\begin{table}[htbp]
	\centering
	\caption{
		Comparison of similarities of image and corresponding point features learned w/o and w/ learnable transformation alignment (LTA). We measure the mean and standard value of similarities on the validation set of SemanticKITTI.}
	\label{table:similarity_statistics}
	\begin{tabular}{l|c|c}
		\toprule[1.2pt]
		& w/o LTA & w/ LTA \\
		\midrule
		Similarity & 0.55 $\pm$ 0.17 & \bf0.69 $\pm$ \bf0.13 \\
		\bottomrule[1.2pt]
	\end{tabular}
\end{table}
\setlength{\tabcolsep}{8pt}

\begin{figure*}[htp]
	\centering
	\includegraphics[width=\textwidth]{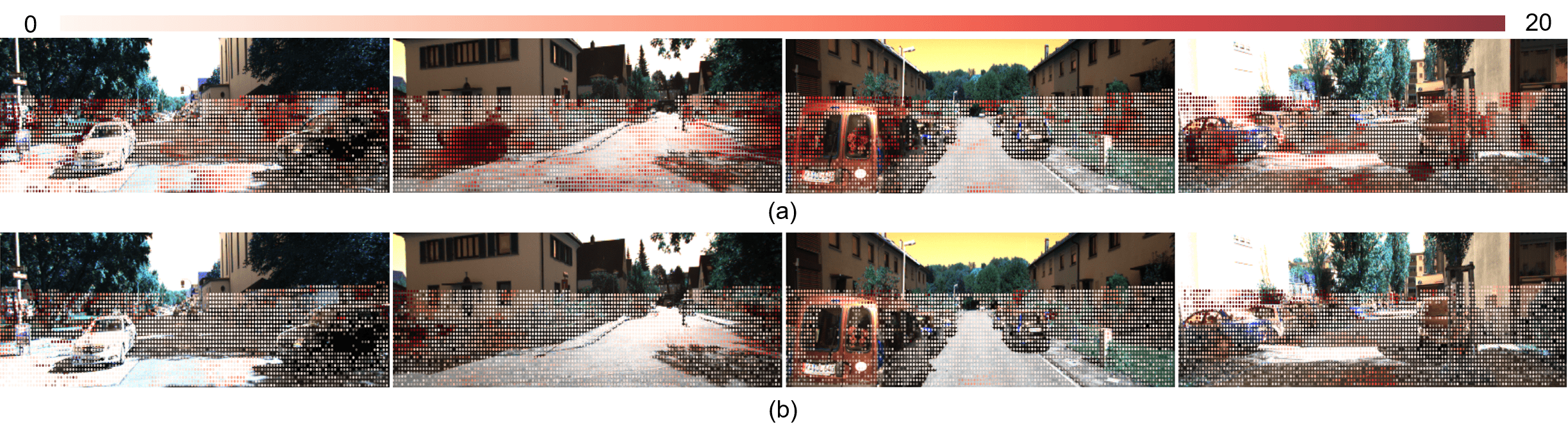} 
        \vspace{-0.7cm}
	\caption{
		Evaluating the accuracy of feature matching across various techniques. (a) Similarities between image and point features are determined using cosine distance, without the aid of learnable transformation alignment, and correspondences between points and pixels are established through hard assignment.
		(b) In contrast, this approach incorporates learnable transformation alignment for calculating feature similarities and employs soft-matching to establish the relationships between points and pixels.
		The matching accuracy reflects the \textit{discrepancy} between the locations of matched pixels and the actual projections of points on the image plane.
	}
	\label{fig:matching_error_vis}
\end{figure*}

\vspace{0.5em}
\noindent\textbf{Effect of Learnable Transformation Alignment.} 
The gap between the features of the two modalities – point clouds and RGB images – is a significant challenge in computing their similarity for matching purposes (see Fig.~\ref{fig:feature_tsne}). 
By transforming features from both modalities into a common feature space, the learnable transformation $\mathbf{W}_{f}$ helps in aligning these features more effectively. 
We further visualize the similarities between image and point features in Fig.~\ref{fig:sk_3dquery_similarity_vis} and Fig.~\ref{fig:sk_2dquery_similarity_vis}.

We also provide a quantitative comparison of similarities between the features of two modalities in Table~\ref{table:similarity_statistics}. Specifically, we computed the cosine similarity between the features of the points and corresponding pixels with and without linear feature transformation, respectively. 
 We can observe that directly computing the cosine similarities yields smaller values for the corresponding points and pixels. However, applying the learnable feature transformation bridges the gap between the two modalities and brings about more reasonable similarities.

\begin{table}[t]
	\centering
	\caption{Feature matching accuracy on SemanticKITTI dataset. The error indicates the discrepancy between the locations of matched pixels and the actual projections of points on the image plane. ``Cosine" denotes directly computing the cosine distance of image and point features as similarities. ``Learnable" denotes learnable transformation alignment. ``Acc." indicates the percentage of fine matchings with errors less than 5 pixels.}
	\label{table:matching_acc}
	\begin{tabular}{l|cc}
		\toprule[1.2pt]
		Method & Error (pixel) & Acc. (\%) \\
		\midrule
		Cosine + Hard-Match & 4.58 $\pm$ 4.70 & 69.22 \\
		Cosine + Soft-Match & 3.91 $\pm$ 4.06 & 74.27 \\
		Learnable + Hard-Match & 3.08 $\pm$ 2.40 & 83.18 \\
		Learnable + Soft-Match & \textbf{2.34 $\pm$ 1.05} & \textbf{90.28} \\
		\bottomrule[1.2pt]
	\end{tabular}
\end{table}

\subsection{Feature Matching Accuracy}\label{sec:feature_matching_acc}

We present a comparative analysis of feature matching accuracy using various feature alignment and point-to-pixel matching methods. The mean and standard deviation of error distances are detailed in Table~\ref{table:matching_acc}. To assess the precision of different methods, we calculated the matching accuracy (Acc.), defined as the percentage of fine matchings with errors less than 5 pixels. The results in Table~\ref{table:matching_acc} demonstrate that the integration of learnable transformation alignment and the soft-matching approach enhances matching accuracy on the SemanticKITTI dataset. Consequently, our method facilitates the network's comprehensive understanding of the relationship between image and point cloud data, learning effective representation to process information from each domain.

Additional visualization of feature matching is presented in Fig.~\ref{fig:matching_error_vis}. Within the overlapping region, we determine the corresponding 2D pixel for each point and evaluate the error. This is done by initially projecting the point onto the image space and then calculating the Euclidean distance between the projected point and its matched 2D pixel. It is evident from our observations that the implementation of our proposed learnable transformation alignment combined with soft-matching markedly reduces the feature-matching error, in comparison to approaches that do not utilize these methods. Notably, in most instances, the errors are confined to within a 3-pixel range.

\begin{figure}[t]
	\centering
	\includegraphics[width=0.5\textwidth]{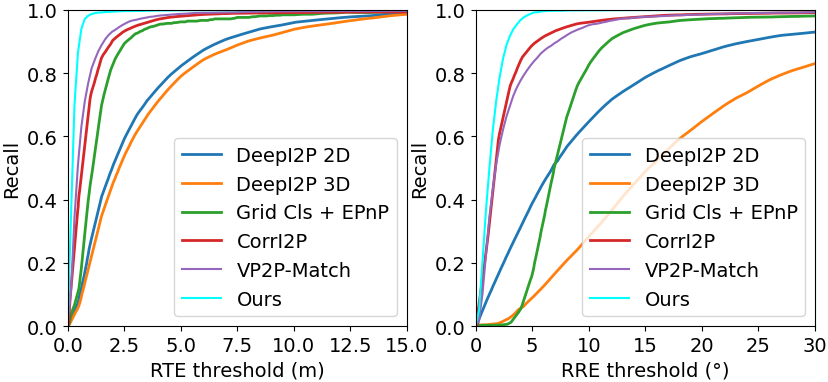} 
	\caption{
		\revise{Illustration of the registration recall performance of various methods against different Relative Translation Error (RTE) and Relative Rotation Error (RRE) thresholds on the SemanticKITTI datasets. The $x$-axis represents the threshold levels for RTE and RRE, while the $y$-axis indicates the recall rate, reflecting the proportion of samples where RREs or RTEs remain below the corresponding threshold.}
	}
	\label{fig:registration_performance}
\end{figure}

\setlength{\tabcolsep}{3pt}
\begin{table}[h]
\centering
    \caption{\revise{Comparison of registration accuracy (mean ± std) achieved by various methods on the SemanticKITTI and nuScenes datasets. The metric ``RTE" measured in meters and ``RRE" measured in degrees are used for evaluation. The best results are highlighted in bold.}}
    \label{table:registry_acc}
    \scalebox{0.92}{
    \begin{tabular}{l|c|c|c|c}
        \toprule
        \multirow{2}{0.1\textwidth}{Method} & \multicolumn{2}{c|}{SemanticKITTI} &  \multicolumn{2}{c}{nuScenes} \\
        \cline{2-5}
        & RTE $\downarrow$ (m) & RRE $\downarrow$ ($^{\circ}$) & RTE $\downarrow$ (m) & RRE $\downarrow$ ($^{\circ}$) \\
        \hline
        BANet~\cite{aich2021bidirectional}+ICP & $3.47\pm 1.44$ & $5.21\pm 3.76$ & $4.09\pm2.28$ & $7.85\pm 3.62$ \\
        Grid Cls. + EPnP ~\cite{li2021deepi2p} & $1.07\pm 0.61$ & $6.48\pm1.66$ & $2.35\pm1.12$ & $7.20\pm1.65$ \\
        DeepI2P (3D)~\cite{li2021deepi2p}      & $1.27\pm0.80$ & $6.26\pm2.29$ & $2.00\pm1.08$ & $7.18\pm1.92$ \\
        DeepI2P (2D)~\cite{li2021deepi2p}      & $1.46\pm0.96$ & $4.27\pm2.74$ & $2.19\pm1.16$ & $3.54\pm2.51$ \\
        EFGHNe~\cite{jeon2022efghnet}          & $3.19\pm1.13$ & $4.95\pm2.50$ & $3.92\pm1.49$ & $5.74\pm3.44$ \\
        CorrI2P~\cite{ren2022corri2p}          & $0.74\pm0.65$ & $2.07\pm1.64$ & $1.83\pm1.06$ & $2.65\pm1.93$ \\
        VP2P~\cite{Zhou2023VP2P}               & $0.67\pm0.48$ & $2.43\pm4.47$ & $0.89\pm1.44$ & $2.15\pm7.03$ \\
        \hline
        Ours                                   & $\pmb{0.27\pm0.26}$ & $\pmb{1.45\pm1.04}$ & $\pmb{0.72\pm0.63}$ & $\pmb{1.25\pm0.93}$ \\
        \bottomrule
    \end{tabular}
    }
\end{table}
\setlength{\tabcolsep}{8pt}


\begin{figure*}[t]
	\centering
	\includegraphics[width=\textwidth]{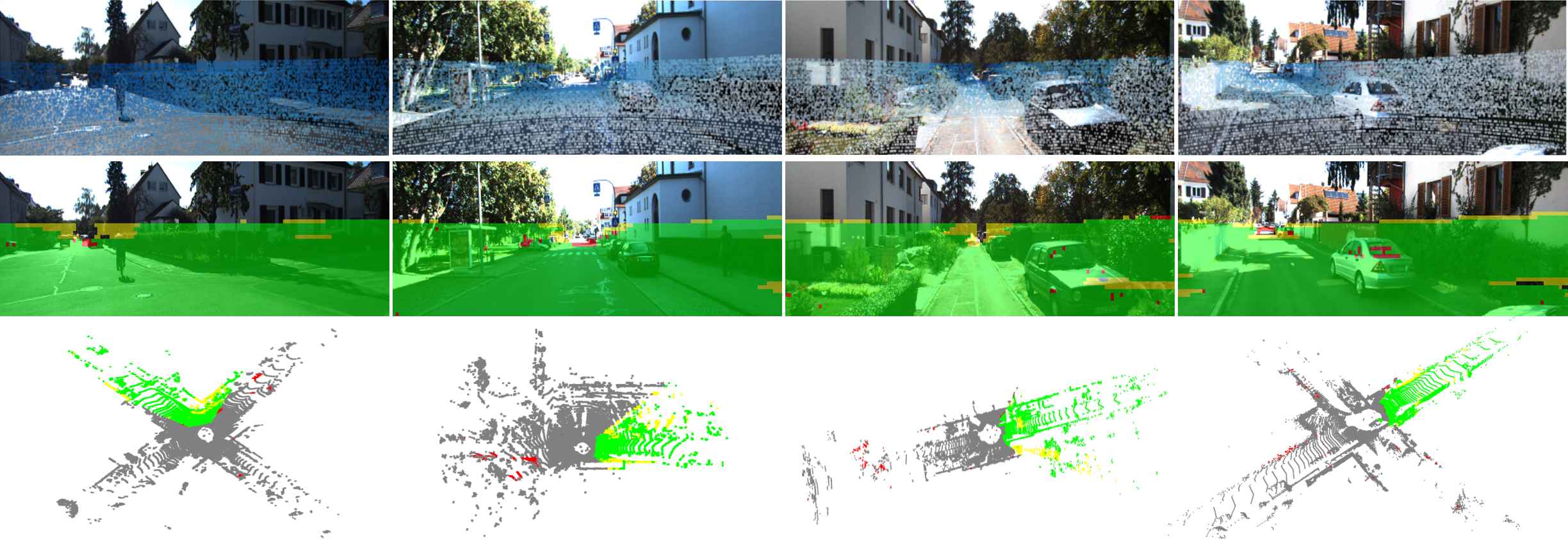} 
	\caption{
		\minor{Visual results of detecting overlapping areas between images and point clouds. The first row depicts the actual projection of the point cloud onto the image plane, with varying colors indicating the depth of the points. The detection results on the image and point cloud are shown in the second and third rows, respectively. In these rows, green pixels and points represent correct detections within the overlapping regions of the image and point cloud. Conversely, red indicates the pixels and points are incorrectly classified in the overlapping areas, and yellow highlights areas are incorrectly identified as non-overlapping.}
	}
	\label{fig:overlapping_area_estimation_vis}
\end{figure*}

\begin{figure}[t]
	\centering
	\includegraphics[width=0.48\textwidth]{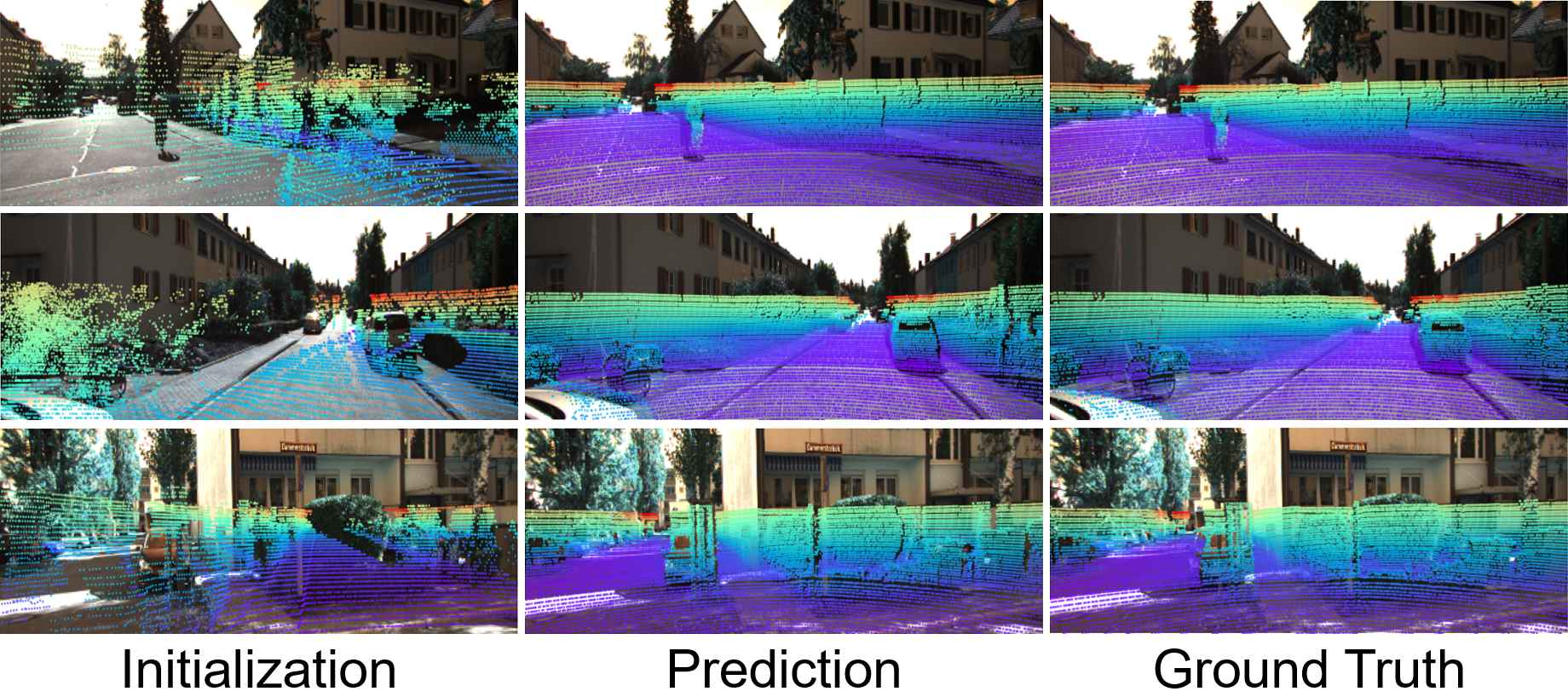} 
	\caption{
		The visual results from the 2D-3D neural calibration on the SemanticKITTI dataset. 
  In the first columns, we show the serious misalignment between the image and point cloud, which has initially undergone random rotation and translation. In the second column, we project the point cloud on the image plane with our estimated spatial pose aligning the camera and LiDAR systems. Meanwhile, the third column illustrates the projections of the point cloud with ground-truth \revise{rigid pose} (camera pose). The painted color indicates the depth of each LiDAR point.
	}
	\label{fig:registration_results_vis}
\end{figure}

\subsection{Accuracy of Image-to-Point Cloud Calibration}
We also compare our approach with other leading methods for image-to-point cloud registration/calibration~\cite{li2021deepi2p,ren2022corri2p,Zhou2023VP2P}. Following standard practices in this field, we used the SemanticKITTI dataset to measure the accuracy of our method, employing metrics such as Relative Translational Error (RTE) and Relative Rotational Error (RRE)~\cite{ren2022corri2p}. We provide a detailed analysis by illustrating the registration recall at various RTE and RRE thresholds on the SemanticKITTI dataset, as shown in Fig.~\ref{fig:registration_performance}. The results clearly indicate the enhanced performance of our technique in terms of registration accuracy.

\revise{We also report the quantitative results presented in Table~\ref{table:registry_acc} to highlight the effectiveness of our proposed method in 2D-3D calibration across the SemanticKITTI and nuScenes datasets. Specifically, our method outperforms all baseline approaches by achieving the lowest RTE and RRE on both datasets.
On the SemanticKITTI dataset, our method achieves an RTE of 0.27 ± 0.26 meters and an RRE of 1.45 ± 1.04 degrees, representing substantial improvements over the competing methods. Notably, the closest baseline, VP2P, reports an RTE of 0.67 ± 0.48 meters and an RRE of 2.43 ± 4.47 degrees, but our approach cuts down the translation error by over 50\% while reducing the rotation error by nearly 40\%. On the nuScenes dataset, our method maintains its superior performance with an RTE of 0.72 ± 0.63 meters and an RRE of 1.25 ± 0.93 degrees.
The consistent reduction in both RTE and RRE demonstrates our model’s robust ability to accurately estimate the rigid pose between the camera and LiDAR systems across different datasets and scenarios.
}

While numerous studies have addressed the task of image-to-point cloud registration, it is noteworthy that these methods predominantly utilize simulated data. In these simulations, point clouds are subjected to random rotation and translation transformations. Consequently, the transformed point clouds are not aligned with the corresponding images. However, the transformations applied in these simulations are idealized, restricted to limited axes and not reflective of the general camera poses found in realistic scenes. This limitation potentially reduces the practical applicability of such methods in real-world scenarios, such as robot localization. In our research, we broaden the scope of image-to-point cloud registration/calibration by extending its application to self-supervised pre-training for 3D perception tasks.


\subsection{Further Discussions}\label{sec:discussions}

Fig.~\ref{fig:overlapping_area_estimation_vis} demonstrates the precise detection of overlapping areas in both image and point cloud domains. In the image domain, inaccuracies are mostly confined to boundary areas. On the side, Fig.~\ref{fig:registration_results_vis} illustrates the registration performance on the SemanticKITTI dataset. The figures in first column reveal noticeable misalignments between the image and the point cloud initialized with random rotation and translation.
Despite the challenging initial conditions, our method accurately estimates the applied \revise{rigid pose}. This precise detection of overlapping areas and accurate estimation of the \revise{rigid pose} indicate that the network effectively understands the spatial information and relationships between each domain. Such insights help explain why the representations learned in the pretext task are advantageous for downstream tasks.

\section{Conclusion}\label{sec:conclusion}
We have introduced an innovative perspective to self-supervised learning by achieving a thorough alignment between two distinct modalities.
In the pretext task of 2D-3D neural calibration, this network not only learns fine-grained matching from individual points to pixels but also achieves a comprehensive alignment between the image and point cloud data, i.e., understanding \minor{the LiDAR-to-camera extrinsic parameters}.
The integration of a soft-matching strategy alongside a differentiable PnP solver makes our framework end-to-end differentiable, thereby facilitating a more comprehensive understanding of multi-modal data.
Besides, our method overcomes the limitations of traditional contrastive learning by introducing a transformation alignment technique that effectively mitigates the domain gap between image and point cloud data. 
The efficacy of our proposed pre-training method is substantiated through extensive experiments across various datasets and fine-tuning tasks, including LiDAR-based 3D semantic segmentation, object detection, and panoptic segmentation. Our results indicate a notable superiority over existing methods, underscoring the potential of our approach in enhancing 3D perception tasks.



%

\appendices


\ifCLASSOPTIONcompsoc
\else
\fi


\ifCLASSOPTIONcaptionsoff
  \newpage
\fi



\bibliographystyle{IEEEtran}
\bibliography{SSL_pretrain}

\end{document}


\title{Self-supervised Learning of LiDAR 3D Point Clouds via 2D-3D Neural Calibration \\ (\textit{Supplementary Materials})}


%
%
%
%

\author{Yifan Zhang,  
	Junhui Hou,~\IEEEmembership{Senior Member,~IEEE},
        Siyu Ren,
	Jinjian Wu,~\IEEEmembership{Senior Member,~IEEE},
        Yixuan Yuan,~\IEEEmembership{Senior Member,~IEEE},
	and Guangming Shi,~\IEEEmembership{Fellow,~IEEE}
	\IEEEcompsocitemizethanks{\IEEEcompsocthanksitem Y. Zhang is with the School of Mechatronic Engineering and Automation, Shanghai University, Shanghai, China, and also with the Department of Computer Science, City University of Hong Kong, Hong Kong. E-mail: yfzhang@shu.edu.cn;
\IEEEcompsocthanksitem J. Hou, and S. Ren are with the Department of Computer Science, City University of Hong Kong, Hong Kong. E-mail: siyuren2-c@my.cityu.edu.hk; jh.hou@cityu.edu.hk;
\IEEEcompsocthanksitem J. Wu and G. Shi are with the School of Artificial Intelligence, Xidian University,
Xi’an 710071, China (e-mail: jinjian.wu@mail.xidian.edu.cn; gmshi@xidian.edu.cn)
\IEEEcompsocthanksitem Y. Yuan is with the Department of Electronic Engineering, The Chinese University of Hong Kong, Hong Kong. E-mail: yxyuan@ee.cuhk.edu.hk
\IEEEcompsocthanksitem  
This work was supported in part by the NSFC Excellent Young Scientists Fund 62422118, in part by the Hong Kong RGC under Grants 11219324 and 11219422, and in part by the Hong Kong ITC under Grant ITS/164/23. (\textit{Corresponding author: Junhui Hou})
		\protect \\
		
	}
}

%
%

\markboth{}%
{Shell \MakeLowercase{\textit{et al.}}: Bare Demo of IEEEtran.cls for Computer Society Journals}
\maketitle

\IEEEdisplaynontitleabstractindextext

%
\IEEEpeerreviewmaketitle
%
In this supplementary material, we provide more experimental results and discussions.

\section{Additional Experimental Results}
\subsection{Analysis of Convergence and Sample Efficiency}

Whether the benefits of our method stem primarily from faster convergence or if they also indicate improved sample efficiency. To answer this question, we conducted additional experiments comparing the convergence behavior and final performance of models trained with and without pre-training.

\begin{figure}[t]
	\centering
	\includegraphics[width=0.49\textwidth]{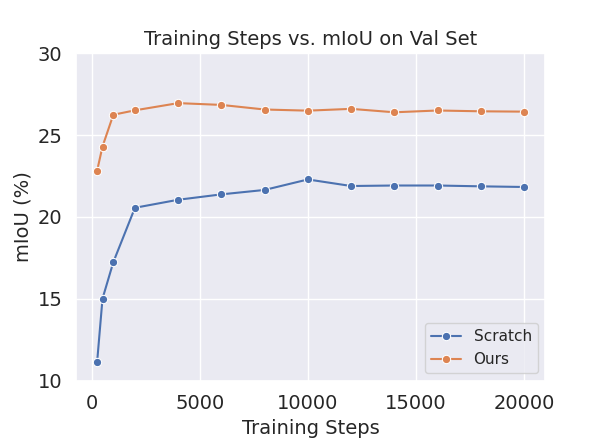} 
	\caption{
		Comparison of our proposed NCLR method with training from scratch using 0.1\% annotated nuScenes data. The results show that our method converges faster and achieves higher semantic segmentation accuracy (mIoU) across various training steps compared to training from scratch.
	}
	\label{fig:convergency}
\end{figure}

Figure~\ref{fig:convergency} illustrates the comparison between our proposed NCLR method and training from scratch using only 0.1\% annotated nuScenes data for the semantic segmentation task. The results show that our pre-trained model not only converges faster but also achieves higher accuracy (mIoU) across different training steps. Importantly, even with an extended training schedule, the model trained from scratch fails to reach the performance level of the pre-trained model. This indicates that the advantages of our pre-training approach extend beyond merely speeding up convergence—it also leads to better sample efficiency and enhanced representation quality.

The experiments confirm that the performance gap is not just due to faster convergence; rather, the pre-trained model consistently provides a superior starting point for downstream fine-tuning, which cannot be compensated by additional training when starting from scratch. The findings validate the effectiveness of our approach in achieving both faster convergence and better final performance, highlighting the dual benefits of pre-training in this context.

\subsection{Results on 2D Semantic Segmentation}
\major{To assess the effectiveness of our method for 2D tasks, we evaluate the performance of the pre-trained 2D backbone on the KITTI validation set using a 2D semantic segmentation task. Specifically, we compare our approach to a baseline using the PSPNet architecture, both pre-trained on ImageNet and with our self-supervised pre-training. The results, as shown in Table~\ref{table:2D_seg_exp}, indicate that the pre-trained 2D backbone in our method outperforms the ImageNet pre-trained model across all categories, demonstrating that the 2D backbone effectively benefits from the 3D features during the pre-training phase. These results validate the potential of our multi-modal learning framework in enhancing performance not only for 3D perception tasks but also for traditional 2D tasks, further emphasizing the utility of leveraging both image and point cloud data in self-supervised learning.}

\begin{table}[htp]
	\centering
	\caption{\major{Comparison of 2D semantic segmentation performance using different pre-training methods on the KITTI validation set.}}
	\label{table:2D_seg_exp}
	\begin{tabular}{l|ccc|l} 
		\Xhline{2\arrayrulewidth}
		\multicolumn{1}{c|}{Method}        & Car   & Pedestrian  & Cyclist  & mIoU   \\ 
		\hline
		PSPNet~\cite{zhao2017pyramid}     & 76.24 & 29.05 & 22.46 & 42.59  \\
		ImageNet Pre-train                & 77.82 & 31.73 & 26.11 & 45.23  \\
		\hline
		Ours                              & \textbf{78.35} & \textbf{32.94} & \textbf{27.15} & \textbf{46.14}  \\
		\Xhline{2\arrayrulewidth}
	\end{tabular}
\end{table}

\setlength{\tabcolsep}{8pt}
\renewcommand{\arraystretch}{1.1}
\begin{table*}[htbp]
	\centering
	\caption{
		Results of repeated experiments using the SECOND detector fine-tuned on the KITTI3D dataset with different random seeds.
	}
	\label{table:3dod_kitti3d_source}
	\scalebox{1.0}{
		\begin{tabular}{l|ccc|ccc|ccc|c} 
			\toprule
			\multirow{2}{*}{\makecell{Pre-training \\Schedule}} & \multicolumn{3}{c|}{Car}        & \multicolumn{3}{c|}{Pedestrian} & \multicolumn{3}{c|}{Cyclist}   & \multirow{2}{*}{Overall}  \\
			~ & Easy  & Moderate & Hard & Easy  & Moderate & Hard & Easy  & Moderate & Hard &                       \\ 
			\midrule
			Scratch   & 90.20    & 81.50 & 78.61 & 53.89   & 48.82    & 44.56 & 82.59   & 65.72    & 62.99 & 67.65 \\
			\midrule
			Ours (Seed-1) & 90.29 & 82.19 & 79.29 & 58.49 & 56.05 & 50.69 & 83.53 & 69.50 & 64.81 & 70.54 \\
			Ours (Seed-2) & 90.13 & 81.77 & 78.83 & 60.09 & 54.12 & 48.67 & 84.77 & 70.76 & 65.88 & \textbf{70.56} \\
			Ours (Seed-3) & 90.23 & 81.99 & 79.05 & 59.20 & 54.75 & 49.32 & 83.64 & 70.16 & 65.13 & 70.38 \\
			\bottomrule
		\end{tabular}
	}
\end{table*}
\renewcommand{\arraystretch}{1.}

\renewcommand{\arraystretch}{1.1}
\begin{table*}[htbp]
	\centering
	\caption{Details of the individual runs for semantic segmentation on the nuScenes dataset. We report the mIoU (\%) on the official validation set for each of the five independent runs.}
	\label{table:semantic_seg_nuscenes_details}
	\scalebox{1.0}{
		\begin{tabular}{cll|ccccc|c}
			\toprule
			\% & Backbone & Method          & \multicolumn{5}{c|}{Runs} & Average and std \\
			\midrule
			0.1\% & MinkUNet &No pre-training             & 21.88 & 21.21 & 22.05 & 21.08 & 21.96 & 21.64 $\pm$0.45\\
			&&PointContrast~\cite{xie2020pointcontrast}   & 26.39   & 27.35 & 27.82 & 26.95 & 26.89 & \textbf{27.08 $\pm$0.54} \\
			&&DepthContrast~\cite{zhang2021depthcontrast} & 21.89 & 21.88 & 21.63 & 21.87 & 21.27 & 21.71 $\pm$0.27\\
			&&ALSO~\cite{boulch2023ALSO}                  & 26.62 & 26.86 & 25.99 & 25.59 & 26.08 & 26.23 $\pm$0.51 \\
			&& Ours                                       & 27.19 & 26.14 & 26.23 & 26.59 & 26.98 & 26.63 $\pm$0.46 \\
			\cmidrule{2-9}
			&SPVCNN & No pre-training                     & 21.97 & 22.30 & 22.09 & 22.18 & 22.45 & 22.20 $\pm$0.19\\
			&& ALSO~\cite{boulch2023ALSO}                 & 24.40 & 24.16 & 25.86 & 25.73 & 23.93 & 24.82 $\pm$0.91 \\
			&& Ours                                       & 26.17 & 25.54 & 26.25 & 26.11 & 24.96 & \textbf{25.81 $\pm$0.55} \\
			\midrule\midrule
			1\% & MinkUNet & No pre-training                        & 34.86 & 35.09 & 34.72 & 34.72 & 35.55 & 34.99 $\pm$0.35\\
			&& PointContrast~\cite{xie2020pointcontrast}    & 37.24 & 37.24 & 36.25 & 36.76 & 37.36 & 36.97 $\pm$0.46\\
			&& DepthContrast~\cite{zhang2021depthcontrast}  & 34.51 & 34.74 & 35.38 & 34.23 & 34.07 & 34.59 $\pm$0.51\\
			&& ALSO~\cite{boulch2023ALSO}                   & 37.42 & 37.52 & 37.15 & 37.11 & 37.94 & 37.43 $\pm$0.34\\
			&& Ours & 37.96 & 37.61 & 38.45 & 36.95 & 38.01 & \textbf{37.80 $\pm$0.56} \\
			\cmidrule{2-9}
			&SPVCNN  & No pre-training                      & 34.27 & 34.94 & 34.26 & 34.10 & 34.37 & 34.39 $\pm$0.32\\
			&& ALSO~\cite{boulch2023ALSO}                   & 37.24 & 37.14 & 37.55 & 37.24 & 37.74 & 37.38 $\pm$0.25\\
			&& Ours & 37.99 & 37.91 & 37.32 & 38.07 & 37.89 & \textbf{37.84 $\pm$0.30} \\
			\midrule\midrule
			10\% & MinkUNet & No pre-training                       & 57.62 & 57.66 & 57.31 & 56.70 & 57.19 & 57.30 $\pm$0.39\\
			&&PointContrast~\cite{xie2020pointcontrast}         & 59.00     & 58.73 & 58.66 & 58.96 & 59.05 & 58.88 $\pm$0.17\\
			&&DepthContrast~\cite{zhang2021depthcontrast}   & 58.03 & 57.00 & 57.36 & 57.56 & 56.90 & 57.37 $\pm$0.46\\
			&&ALSO~\cite{boulch2023ALSO}                    & 58.63 & 58.62 & 59.11 & 59.28 & 59.35 & 59.00 $\pm$0.35\\
			&& Ours & 59.79 & 58.76 & 59.35 & 59.69 & 59.72 & \textbf{59.46 $\pm$0.43} \\
			\cmidrule{2-9}
			&SPVCNN & No pre-training                       & 57.37 & 56.97 & 57.34 & 56.75 & 57.18 & 57.12 $\pm$0.26\\
			&& ALSO~\cite{boulch2023ALSO}                   & 58.15 & 58.56 & 58.42 & 58.48 & 58.60 & 58.44 $\pm$0.18\\
			&& Ours & 58.94 & 59.66 & 59.48 & 58.86 & 58.99 & \textbf{59.19 $\pm$0.36} \\
			\midrule\midrule
			50\% & MinkUNet & No pre-training                       & 68.80 & 68.90 & 68.94 & 69.31 & 69.01 & 68.99 $\pm$0.19\\
			&&PointContrast~\cite{xie2020pointcontrast}     & 69.15 & 69.09 & 69.39 & 69.42 & 69.75 & 69.36 $\pm$0.26\\
			&&DepthContrast~\cite{zhang2021depthcontrast}   & 69.12 & 69.04 & 69.38 & 69.57 & 68.66 & 69.15 $\pm$0.35\\
			&&ALSO~\cite{boulch2023ALSO}                    & 69.69 & 69.58 & 69.93 & 69.66 & 70.17 & 69.81 $\pm$0.24\\
			&& Ours & 71.34 & 71.36 & 71.20 & 71.11 & 71.10 & \textbf{71.22 $\pm$0.12} \\
			\cmidrule{2-9}
			&SPVCNN & No pre-training                       & 69.24 & 69.06 & 68.68 & 68.74 & 69.09 & 68.96 $\pm$0.24\\
			&& ALSO~\cite{boulch2023ALSO}                   & 69.55 & 69.77 & 69.47 & 69.24 & 69.65 & 69.54 $\pm$0.20\\
			&& Ours & 70.91 & 71.17 & 71.09 & 71.27 & 70.78 & \textbf{71.04 $\pm$0.20} \\
			\midrule\midrule
			100 \% & MinkUNet & No pre-training             & 71.21 & 71.35 & 71.20 & 70.93 & 71.32 & 71.20 $\pm$0.17\\
			&&PointContrast~\cite{xie2020pointcontrast}     & 71.12 & 71.27 & 70.90 & 70.94 & 71.31 & 71.11 $\pm$0.19\\
			&&DepthContrast~\cite{zhang2021depthcontrast}   & 71.31 & 71.20 & 71.30 & 70.81 & 71.36 & 71.20 $\pm$0.22\\
			&&ALSO~\cite{boulch2023ALSO}                    & 71.95 & 71.92 & 71.60 & 71.88 & 71.39 & 71.75 $\pm$0.24\\
			&& Ours & 73.04 & 72.58 & 72.73 & 72.45 & 72.60 & \textbf{72.68 $\pm$0.22} \\
			\cmidrule{2-9}
			&SPVCNN & No pre-training                       & 70.82 & 70.79 & 70.56 & 70.86 & 70.41 & 70.69 $\pm$0.19\\
			&& ALSO~\cite{boulch2023ALSO}                   & 71.41 & 71.18 & 70.99 & 71.20 & 71.48 & 71.25 $\pm$0.20\\
			&& Ours & 73.35 & 73.25 & 72.83 & 72.68 & 73.06 & \textbf{73.03 $\pm$0.28} \\
			\bottomrule
		\end{tabular}
	}
\end{table*}
\renewcommand{\arraystretch}{1.}

\renewcommand{\arraystretch}{1.05}
\begin{table*}[htbp]
	\centering
	\caption{Details of the individual runs for semantic segmentation on the SemanticKITTI dataset. We report the mIoU (\%) on the official validation set for each of the five independent runs.}
	\label{table:semantic_seg_sk_details}
	\scalebox{1.0}{
		\begin{tabular}{cll|ccccc|c}
			\toprule
			\% & Backbone & Method          &  \multicolumn{5}{c|}{Runs} & Average and std \\
			\midrule
			0.1\% & MinkUNet &No pre-training          & 30.22 & 29.99 & 29.74 & 30.15 & 29.77 & 29.97 $\pm$0.22\\
			&&PointContrast~\cite{xie2020pointcontrast}  & 32.79	& 31.84	& 31.88	& 32.96 & 32.60 & 32.41 $\pm$0.52\\
			&&DepthContrast~\cite{zhang2021depthcontrast}   & 32.43	& 32.09	& 33.01 & 32.24 & 32.80 & 32.51 $\pm$0.38\\
			&&SegContrast~\cite{nunes2022segcontrast}    & 32.65 & 32.38 & 32.48 & 32.18 & 31.83 & 32.30	$\pm$0.31\\
			&& ALSO~\cite{boulch2023ALSO}       & 34.97 & 34.83 & 34.81 & 35.10 & 35.11 & 34.96 $\pm$0.14\\
			&& Ours & 39.29 &  39.65 & 38.97 & 39.53 & 38.77 & \textbf{39.24 $\pm$ 0.37} \\
			
			\cmidrule{2-9}
			&SPVCNN &No pre-training   & 30.94 & 30.81 & 30.66 & 30.47 & 30.81 & 30.74 $\pm$0.18 \\
			&& ALSO~\cite{boulch2023ALSO}    & 35.35 & 34.78 & 34.71 & 34.93 & 35.43 & 35.04 $\pm$0.33 \\
			&& Ours & 39.15 &  38.96 & 38.83 & 38.34 & 38.53 & \textbf{38.76 $\pm$ 0.33} \\
			
			\midrule\midrule
			1\% & MinkUNet &No pre-training          & 45.1	& 46.32	& 46.59	& 46.68	& 46.49 & 46.24	$\pm$0.65 \\
			&&PointContrast~\cite{xie2020pointcontrast}   	& 47.71	& 47.97	& 48.22	& 47.25	& 48.43 & 47.92	$\pm$0.46 \\
			&&DepthContrast~\cite{zhang2021depthcontrast}   	& 49.62	& 49.12	& 48.59	& 48.94	& 48.74 & 49.00 $\pm$0.40 \\
			&&SegContrast~\cite{nunes2022segcontrast}    	& 48.91	& 49.35	& 48.87	& 48.81	& 48.54 & 48.90 $\pm$0.29 \\
			&& ALSO~\cite{boulch2023ALSO}         & 50.04	& 50.28	& 49.43	& 50.41	& 50.02	& 50.04 $\pm$0.38 \\
			&& Ours & 53.45 &  53.49 & 54.21 & 53.16 & 53.49 & \textbf{53.56 $\pm$ 0.39} \\
			
			\cmidrule{2-9}
			&SPVCNN &No pre-training    &46.24 & 47.23 & 46.53 & 46.26 & 46.67 & 46.59 $\pm$0.40\\
			&& ALSO~\cite{boulch2023ALSO}    & 49.34 & 49.75 & 49.05 & 48.90 & 48.32 & 49.07 $\pm$0.53 \\
			&& Ours & 52.90 &  52.43 & 52.56 & 52.81 & 52.70 & \textbf{52.68 $\pm$ 0.19} \\
			
			\midrule\midrule
			10\% & MinkUNet &No pre-training          & 57.04 & 58.74 & 57.71 & 56.27 & 58.01 & 57.55 $\pm$0.94\\
			&&PointContrast~\cite{xie2020pointcontrast}    & 59.48 & 59.78 & 60.44 & 59.26 & 59.56 & 59.70 $\pm$0.45\\
			&&DepthContrast~\cite{zhang2021depthcontrast}    & 59.49 & 60.74 & 60.27 & 60.46 & 60.75 & 60.34 $\pm$0.52\\
			&&SegContrast~\cite{nunes2022segcontrast}      & 59.63 & 58.57 & 58.45 & 58.78 & 58.29 & 58.74 $\pm$0.53\\
			&& ALSO~\cite{boulch2023ALSO}        & 60.41 & 60.45 & 60.47 & 60.54 & 60.43 & 60.46 $\pm$0.05\\
			&& Ours & 60.99 &  61.81 & 61.59 & 61.59 & 60.88 & \textbf{61.37 $\pm$ 0.41} \\
			
			\cmidrule{2-9}
			&SPVCNN &No pre-training    &58.8 & 58.95 & 59.21 & 59.47 & 57.85 & 58.86 $\pm$0.62 \\
			&& ALSO~\cite{boulch2023ALSO}    & 60.71 & 60.32 & 60.97 & 60.32 & 60.66 & 60.60 $\pm$0.28 \\
			&& Ours & 60.55 &  61.15 & 60.65 & 61.62 & 61.39 & \textbf{61.07 $\pm$ 0.46} \\
			
			\midrule\midrule
			50\% & MinkUNet &No pre-training         & 61.48 & 62.33 & 61.88 & 61.80 & 61.31 & 61.76 $\pm$0.39\\
			&&PointContrast~\cite{xie2020pointcontrast}   & 62.68 & 62.91 & 62.54 & 62.35 & 63.19	& 62.73 $\pm$0.33\\
			&&DepthContrast~\cite{zhang2021depthcontrast}   & 63.24 & 63.31 & 63.16 & 62.44 & 62.37	& 62.90 $\pm$0.46\\
			&&SegContrast~\cite{nunes2022segcontrast}     & 62.58 & 62.20 & 61.61 & 61.74 & 62.46 & 62.12 $\pm$0.43\\
			&& ALSO~\cite{boulch2023ALSO}          & 63.09 & 63.43 & 62.99 & 63.28 & 64.15 & 63.39 $\pm$0.46\\
			&& Ours & 63.88 &  63.12 & 63.37 & 63.82 & 63.51 & \textbf{63.54 $\pm$ 0.32} \\
			
			\cmidrule{2-9}
			&SPVCNN &No pre-training    &61.32 & 62.15 & 61.61 & 61.7 & 62.39 & 61.83 $\pm$0.43 \\
			&& ALSO~\cite{boulch2023ALSO}    & 63.4 & 63.4 & 63.45 & 64.08 & 63.44 & 63.55 $\pm$0.29 \\
			&& Ours & 63.71 &  64.26 & 64.12 & 63.97 & 63.89 & \textbf{63.99 $\pm$ 0.21} \\
			
			\midrule\midrule
			100\% & MinkUNet &No pre-training          & 62.49 & 62.35 & 62.98 & 62.50 & 63.06	& 62.68 $\pm$0.32\\
			&&PointContrast~\cite{xie2020pointcontrast}    & 63.57 & 63.14 & 63.13 & 63.95 & 63.26 & 63.41 $\pm$0.35\\
			&&DepthContrast~\cite{zhang2021depthcontrast}    & 63.76 & 64.31 & 63.52 & 63.54 & 64.12 & 63.85 $\pm$0.35\\
			&&SegContrast~\cite{nunes2022segcontrast}      & 62.64 & 61.57 & 62.53 & 62.24 & 62.45 & 62.29 $\pm$0.43\\
			&& ALSO~\cite{boulch2023ALSO}          & 64.29 & 63.75 & 63.75 & 63.34 & 63.07 & 63.64 $\pm$0.46\\
			&& Ours & 63.79 &  63.88 & 63.92 & 64.38 & 63.59 & \textbf{63.91 $\pm$ 0.29} \\
			
			\cmidrule{2-9}
			&SPVCNN &No pre-training         &62.39 & 62.86 & 62.33 & 62.88 & 62.82 & 62.66 $\pm$0.27 \\
			&& ALSO~\cite{boulch2023ALSO}    & 63.60 & 64.04 & 63.59 & 63.93 & 63.76 & 63.78 $\pm$0.20 \\
			&& Ours & 63.97 &  64.24 & 64.21 & 63.72 & 64.22 & \textbf{64.07 $\pm$ 0.23} \\
			\bottomrule
		\end{tabular}
	}
\end{table*}
\renewcommand{\arraystretch}{1.}

\begin{figure*}[t]
    \centering
    \includegraphics[width=0.85\textwidth]{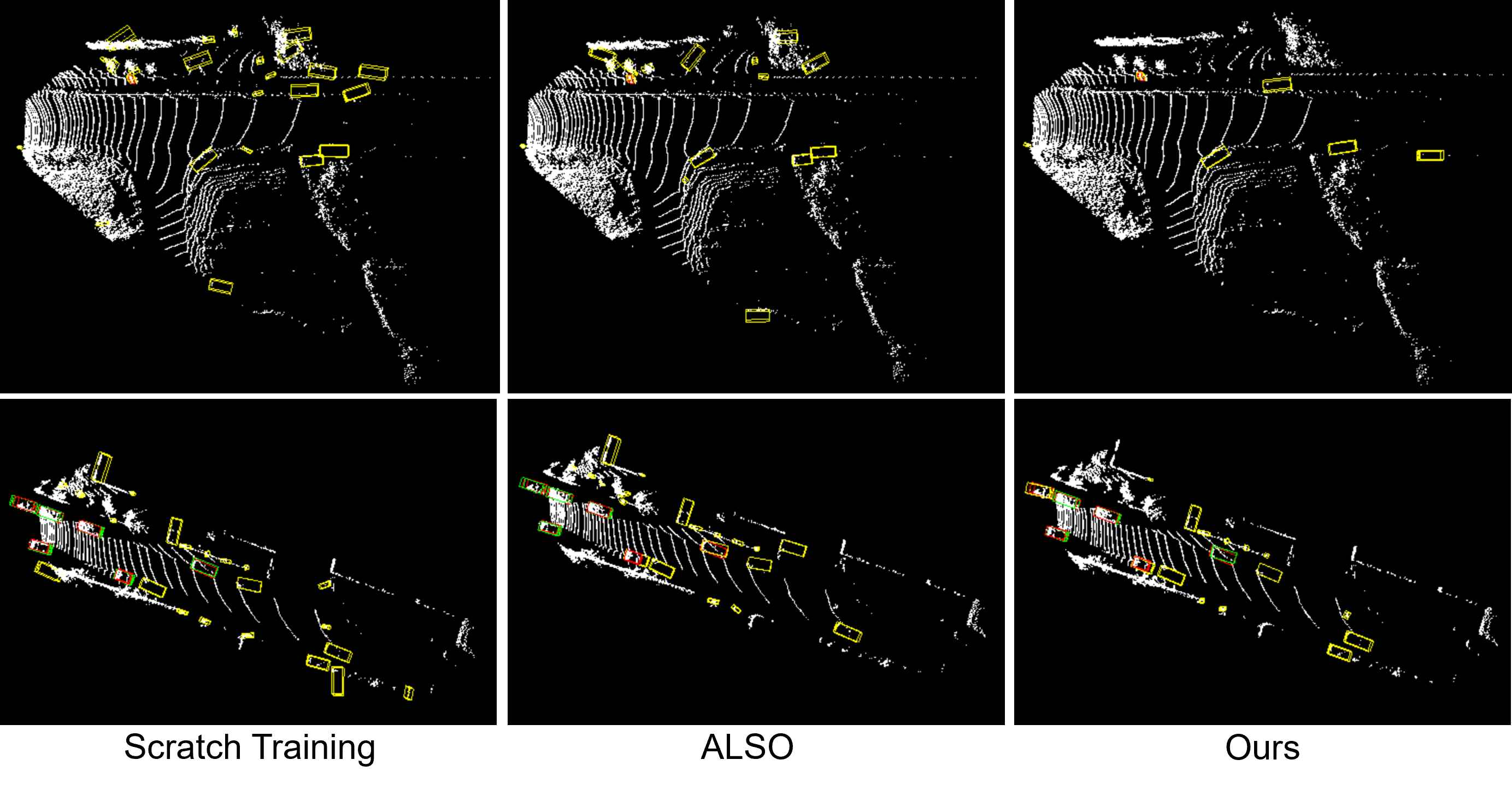} 
    \caption{
        \major{Qualitative comparison of fine-tuned results on the KITTI 3D detection dataset. We compare the performance of scratch training, ALSO~\cite{boulch2023ALSO}, and our method (Ours) on fine-tuned models. We use red, green, and yellow to denote the \textcolor{red}{ground-truth}, \textcolor{green}{true positive} and \textcolor{yellow}{false positive} bounding boxes, respectively.}
    }
    \label{fig:kitti_visualize}
\end{figure*}

\begin{figure*}[htp]
    \centering
    \includegraphics[width=0.99\textwidth]{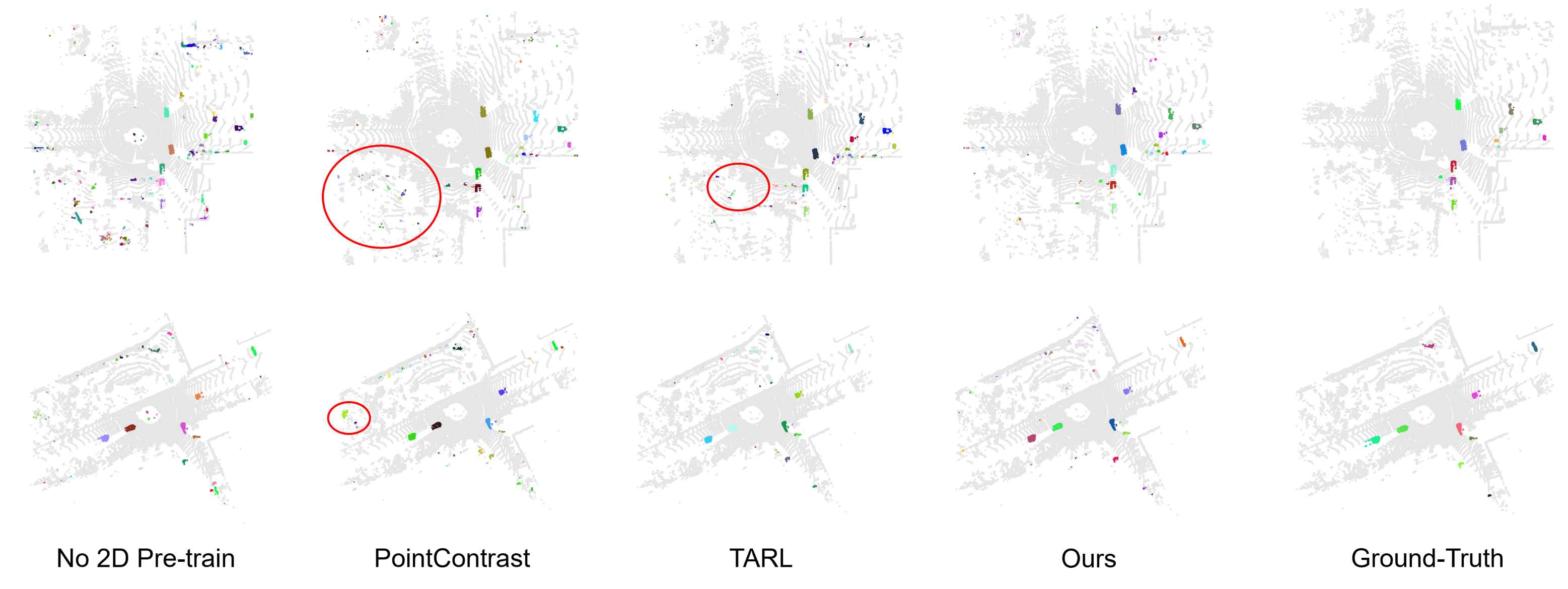} 
    \caption{
        \major{Qualitative comparison of 3D panoptic segmentation results on the SemanticKITTI dataset. The images show the predicted segmentation outputs using different methods: No 2D Pre-train, PointContrast~\cite{xie2020pointcontrast}, TARL~\cite{nunes2023TARL}, Ours, and the Ground-Truth labels. Red circles highlight key differences in performance, where our method demonstrates better segmentation with fewer false positives compared to the other approaches. Best viewed in color and zoom in for more details.}
    }
    \label{fig:panoptic_vis}
\end{figure*}

\subsection{Detailed 3D Detection Results on KITTI3D}
We repeated our fine-tuning experiments on the KITTI3D dataset using the SECOND detector three times with different random seeds. The results are summarized in Table~\ref{table:3dod_kitti3d_source}. 
As shown in the table, even with different seeds, the performance variations are not significant. The consistent improvement compared with the scratch training across different seeds indicates that our improvements are not due to random noise but are indeed robust and repeatable.

\subsection{Detailed 3D Semantic Segmentation Results}
For the semantic segmentation experiments in the main paper, all reported scores are averaged over five independent runs. We have provided detailed results for these runs here: Table~\ref{table:semantic_seg_nuscenes_details} presents the individual scores for the nuScenes experiments, and Table~\ref{table:semantic_seg_sk_details} provides the scores for the SemanticKITTI experiments. 
In each table, we also report the average score (as shown in the main paper) and the corresponding standard deviation.
This additional information reinforces the reliability and reproducibility of our reported results.

\subsection{Effect of the Image Backbone}
\major{To gain a more comprehensive understanding of the adaptability and performance of our method, we evaluate it with different 2D visual feature extraction pre-trained models. Specifically, we conduct additional experiments using several 2D self-supervised learning methods, including OBoW, MoCov2, and DINO, for 2D feature extraction. The results of these experiments, presented in Table~\ref{table:2D_backbone}, show the performance of our method on the SemanticKITTI semantic segmentation task using various pre-training strategies. As indicated, the performance remains consistent across the different pre-training methods, with only marginal improvements observed for DINO. This demonstrates that our approach is robust to different 2D pre-training strategies and performs competitively even without 2D pretraining.}

\setlength{\tabcolsep}{9pt}
\begin{table}[t]
\centering
    \caption{
        \major{Performance comparison of our method on SemanticKITTI semantic segmentation with different 2D pre-training strategies. We evaluate the model with no pre-training as well as with several 2D self-supervised methods. We report the results of fine-tuning with 0.1\% of training data.}
    }
    \label{table:2D_backbone}
    \begin{tabular}{l|cccc} 
    \hline
     & \makecell*[c]{No 2D \\ Pre-training}   & OBoW~\cite{obow2021} & MoCov2~\cite{chen2020mocov2} & DINO~\cite{maxime2023dinov2}  \\
    \hline
    mIoU     & 39.23   &   39.07    & 39.16      & 39.30     \\
    \hline
    \end{tabular}
\end{table}
\setlength{\tabcolsep}{6pt}

\subsection{More Visual Results}
 
\major{The qualitative results shown in Figures~\ref{fig:kitti_visualize} and \ref{fig:panoptic_vis} highlight the effectiveness of our method across two downstream tasks. In Figure~\ref{fig:kitti_visualize}, we present the 3D object detection results on the KITTI validation set, where our approach demonstrates superior performance over scratch training and the ALSO method~\cite{boulch2023ALSO}. Specifically, our method reduces the number of false positives and missed detections, providing more accurate bounding box predictions. Similarly, in Figure~\ref{fig:panoptic_vis}, we show the 3D panoptic segmentation results on the SemanticKITTI dataset. Our method outperforms the other approaches, including PointContrast~\cite{xie2020pointcontrast} and TARL~\cite{nunes2023TARL}, by achieving better segmentation accuracy, as indicated by fewer false positive detections. These visual comparisons highlight the robustness and versatility of our method across different tasks, demonstrating its ability to consistently produce high-quality results.}

\section{Discussion}
\subsection{Insight of Holistic Rigid Pose Estimation}

In this part, we clarify the differences between local matching from points to pixels and the holistic rigid pose estimation.

First, local matching focuses on establishing fine-grained correspondences between points in the LiDAR point cloud and pixels in the image, typically relying on local feature similarities. However, local feature matching alone is \textbf{insufficient} to determine the global alignment or pose between the two coordinate systems. Due to differences in the operational principles and fields of view of the camera and LiDAR sensors, the image and point cloud do\textbf{ not perfectly overlap}. If we mistakenly match points in non-overlapping areas with pixels, incorrect point-pixel correspondences could mislead the optimizer when solving for the pose, resulting in inaccurate global alignment. To address this, we perform overlapping area detection to identify the points and pixels that genuinely correspond to each other before building dense point-pixel correspondences. This step is critical to avoid introducing noise from irrelevant regions and ensures the robustness of the subsequent pose estimation process.

Second, our method goes beyond simply learning discriminative features through contrastive losses for local matching from points to pixels. Once the matched point-pixel correspondences are established, we \textbf{explicitly} estimate the camera pose through a differentiable PnP solver, which aligns the entire coordinate systems. The \textbf{additional} pose estimation loss imposes global consistency and regularization on the learned features, enhancing the network’s ability to capture both local details and complete spatial relationships between the modalities. This explicit pose estimation is vital because it bridges the gap between local feature matching and global scene understanding, allowing the network to produce more reliable and transferable representations for downstream tasks.


%




\ifCLASSOPTIONcompsoc
\else
\fi


\ifCLASSOPTIONcaptionsoff
  \newpage
\fi



\bibliographystyle{IEEEtran}
\bibliography{SSL_pretrain}